\documentclass[a4paper, review]{jds}      

\jdsmonth{}                     
\jdsyear{}                      
\jdsvolume{}                    
\jdsissue{}                     
\jdsdoi{}                       

\usepackage{amsfonts,amsmath,amssymb,amsthm}
\usepackage{booktabs}
\usepackage{url}       
\usepackage{hyperref}  
\usepackage{lipsum}
\usepackage{tikz}
\newtheorem{definition}{Definition}
\newtheorem{proposition}{Proposition}
\usepackage{algorithm}
\usepackage{algpseudocode}
\usepackage{longtable}
\usepackage{hyperref}
\usepackage{soul}

\nolinenumbers
\title[Diffeomorphic Time Warping (DiffTW)]{Time Series Classification through Diffeomorphic Time Warping (DiffTW)}

\author[1]{Vicky Geneva Haney\footnote{Corresponding author. Email: vhaney@pdx.edu}}
\author[2]{Kamel Lahouel\thanks{Corresponding author. Email: klahouel@tgen.org}}
\author[2]{Victor Rielly\footnote{Corresponding author. Email: vrielly@tgen.org}}
\author[1]{Bruno M. Jedynak\footnote{Corresponding author. Email: bjedyna2@pdx.edu}}
\affil[1]{Department of Mathematics and Statistics, Portland State University, USA}
\affil[2]{The Translational Genomics Research Institute (TGen), USA}

\begin{document}

\maketitle
\thispagestyle{plain}
\nolinenumbers
\begin{abstract}
  Time series classification involves learning a mapping from a continuous, temporally ordered sequence of real-valued observations to discrete response variables, like class labels. This task is fundamental in domains, including health monitoring, where temporal structure is critical for prediction. Dynamic Time Warping (DTW) is a standard technique for measuring similarity between sequences varying in time or speed. However, DTW is restricted to discrete point matching. Moving beyond pairwise alignment, we propose a theoretical framework learning mappings between real-valued functions. These mappings approximate the flow associated with the characteristic curves of a linear transport equation with a space-dependent velocity field, providing a diffeomorphic transformation between time series. Using the method of characteristics, we transform this partial differential equation into ordinary differential equations (ODEs) modeling system dynamics. The objective function to learn these ODEs derives from the fundamental theorem of calculus. To enable flexible, expressive representations of the velocity field, we utilize reproducing kernel Hilbert spaces and optimal control methods. Our method, Diffeomorphic Time Warping (DiffTW), provides a theoretically grounded dissimilarity measure. Using a 1-nearest neighbor classifier, DiffTW outperforms unconstrained DTW on 39 of 85 datasets with 3 ties; however, constrained DTW outperforms DiffTW on 48 of 85 datasets with 5 ties.
\end{abstract}

\begin{keywords} 
  Backpropagation;
  DTW;
  Nearest neighbor classification; RKHS
\end{keywords}

\section{Introduction}%
\label{sec:intro}

Dynamic Time Warping (DTW) is a widely used method for comparing time series that may differ in sampling rate or temporal distortion. DTW computes an optimal alignment path based on a cumulative cost matrix, and the resulting alignment cost serves as a dissimilarity measure \citet{dtwcode, cardiac2011}. While DTW is effective, it remains a pairwise alignment method and does not model the underlying process generating the sequences. Many extensions, such as modified distances, regularization, and differentiable variants, \citet{liu2024novel, LIU2024119921, softdtw, cdtw}, still operate within the alignment framework. 

Instead of aligning sequences, we learn the dynamics transforming one signal into another. For processes with continuous temporal deformations, we represent these transformations using a first-order partial differential equations (PDE), enabling smooth maps and a dynamical view of time series.

Using the method of characteristics, we reduce the PDE to a system of ordinary differential equations (ODEs) and learn the associated vector field in a reproducing kernel Hilbert space (RKHS). Our approach is conceptually related to large deformation diffeomorphic metric mapping (LDDMM) \citet{lddm}. Our approach differs in two key aspects: we use a time-independent vector field and an explicit kernel composed of random Fourier features \citet{rahimi2007random}. We call the resulting algorithm Diffeomorphic Time Warping (DiffTW).

We evaluated DiffTW on electrocardiogram (ECG) classification tasks where DTW has been extensively studied \citet{cardiac2011, bakeoff, UCRpaper}. Using standard ECG datasets and subsets of the University of California, Riverside (UCR) Time Series Archive \citet{UCRpaper}, we assessed the performance of DiffTW using a 1-nearest neighbor (1-NN) classifier. To establish our baselines, we compared our DiffTW results against a standard Python implementation of unconstrained DTW \citet{dtwcode}, as well as the constrained and unconstrained DTW accuracies (1-error rate) provided by the UCR archive.

The rest of this paper is structured as follows: Section~\ref{Sec:Methods} describes the methodology we used to learn the vector fields. Section~\ref{Sec:Exp} details the experiments we conducted using this approach. Section~\ref{Sec:Results} provides an overview of our results compared to DTW, and finally, Section~\ref{Sec:Conclusion} summarizes our work and discusses its impact on time series classification.

\section{Methods}\label{Sec:Methods}

\subsection{Defining the Partial Differential Equation}  
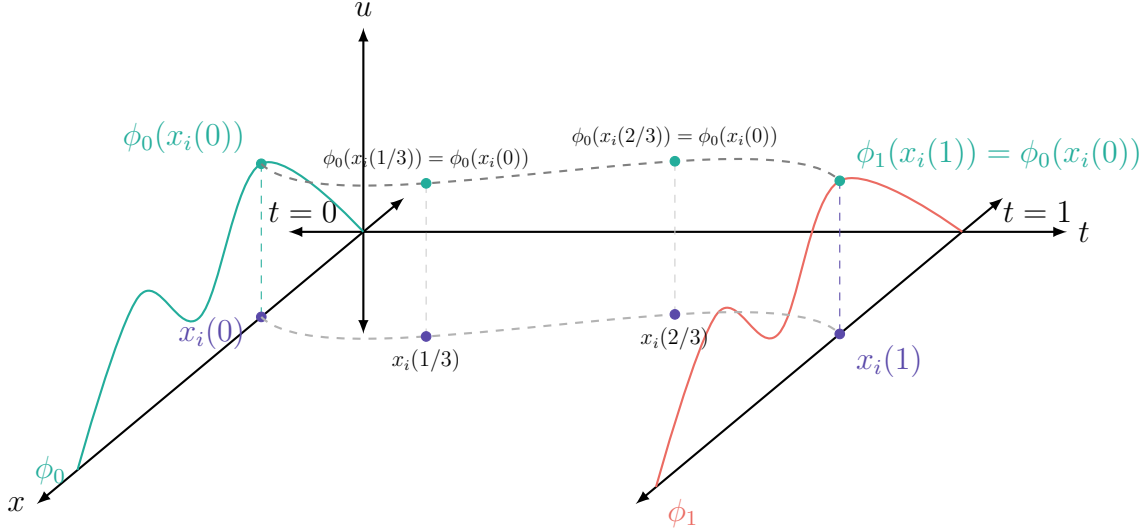
\begin{figure}[!h]
\label{fig:mofc}
\tikzset{
  xy plane/.style={cm={1,0,0,1,(0,0)}},
  xz plane/.style={cm={-0.6,-0.3,0,1,(0,0)}},
  yz plane/.style={cm={1,0,-0.6,-0.3,(0,0)}},
  hand drawn/.style={
    smooth,
    line cap=round,
    line join=round,
  }
}

\definecolor{tealcurve}{RGB}{35, 173, 155}
\definecolor{redcurve}{RGB}{205, 107, 98}
\definecolor{purplechar}{RGB}{124, 73, 169}
\definecolor{bluechar}{RGB}{62, 107, 192}

\begin{tikzpicture}[
    x={(2.2cm, 0cm)},       
    y={(0cm, 1.5cm)},       
    z={(-0.6cm, -0.5cm)},   
    >=latex,                
    scale=0.9
]

\definecolor{myteal}{RGB}{35, 173, 155}
\definecolor{myred}{RGB}{235, 107, 98}
\definecolor{mypurple}{RGB}{90, 73, 169}

\draw[<->, thick] (0, 0, -1) -- (0, 0, 8) node[left] {$x$};
\draw[<->, thick] (-0.5, 0, 0) -- (4.7, 0, 0) node[right] {$t$};
\draw[<->, thick] (0, -1, 0) -- (0, 2, 0) node[above] {$u$};

\draw[<->,thick] (4, 0, -1) -- (4, 0, 8);
\node[right] at (4.2, .2, 0) {$t=1$};
\node[left] at (-.1, .2, 0){$t=0$};

\draw[thick, myteal] plot [smooth, tension=0.6] coordinates {
    (0, 0, 0)
    (0, 1.5, 2.5) 
    (0, 0.5, 4)
    (0, 1.2, 5.5) 
    (0, 0, 7)
} node[left] {$\phi_0$};

\draw[thick, myred] plot [smooth, tension=0.6] coordinates {
    (4, 0, 0)
    (4, 1.5, 3)   
    (4, 0.5, 4.5)
    (4, 1.2, 6)   
    (4, 0, 7.5)
} node[below right] {$\phi_1$};

\draw[dashed, myteal] (0, 1.5, 2.5) -- (0, 0, 2.5);
\filldraw[myteal] (0, 1.5, 2.5) circle (2pt) node[above left, xshift=-2pt] {$\phi_0(x_i(0))$};
\filldraw[mypurple] (0, 0, 2.5) circle (2pt) node[below left, xshift=-2pt, yshift=4pt] {$x_i(0)$};

\draw[dashed, mypurple] (4, 1.5, 3) -- (4, 0, 3);
\filldraw[myteal] (4, 1.5, 3) circle (2pt) node[above right, xshift=2pt] {$\phi_1(x_i(1)) = \phi_0(x_i(0))$};
\filldraw[mypurple] (4, 0, 3) circle (2pt) node[below right, xshift=2pt] {$x_i(1)$};

\draw[dashed, gray, thick] (0, 1.5, 2.5) .. controls (1.0, 1.5, 4.5) and (3.0, 1.5, 1.0) .. (4, 1.5, 3)
    coordinate[pos=0.333] (top1)
    coordinate[pos=0.667] (top2);

\draw[dashed, gray!60, thick] (0, 0, 2.5) .. controls (1.0, 0, 4.5) and (3.0, 0, 1.0) .. (4, 0, 3)
    coordinate[pos=0.333] (bot1)
    coordinate[pos=0.667] (bot2);

\draw[dashed, gray!40] (bot1) -- (top1);
\draw[dashed, gray!40] (bot2) -- (top2);

\filldraw[mypurple] (bot1) circle (2pt);
\filldraw[mypurple] (bot2) circle (2pt);

\filldraw[myteal] (top1) circle (2pt);
\filldraw[myteal] (top2) circle (2pt);

\node[below, scale=0.7, yshift=-3pt] at (bot1) {$x_i(1/3)$};
\node[below, scale=0.7, yshift=-3pt] at (bot2) {$x_i(2/3)$};

\node[above, scale=0.65, yshift=4pt] at (top1) {$\phi_0(x_i(1/3)) = \phi_0(x_i(0))$};
\node[above, scale=0.65, yshift=4pt] at (top2) {$\phi_0(x_i(2/3)) = \phi_0(x_i(0))$};

\end{tikzpicture}
\caption{Visualization of continuous time warping via the method of characteristics. Points $x_i(0)$ from the reference signal are transported along characteristic curves $x_i(t)$ over time $t \in [0,1]$. By construction, the signal value $u$ is conserved along these trajectories, meaning $\phi_0(x_i(t)) = \phi_0(x_i(0))$. The proposed model optimizes the underlying vector field $\alpha(x)$ to construct the specific paths $x_i(t)$ that continuously deform $\phi_0$ into optimal alignment with the target signal $\phi_1$.}
\end{figure}

We aim to transition from measuring similarity between sequences as done in DTW to
modeling the underlying dynamics and framing the alignment process as a continuous physical flow governed by a first-order advection partial differential equation (PDE). Let $u: [0,1] \times [0,1]\mapsto \mathbb{R}$ represent the signal evolving from a source sequence $\phi_0$ to a target sequence $\phi_1$. The boundary states of this evolution are defined as:
\begin{equation*}
    u(x,0) = \phi_0(x), \quad u(x,1) = \phi_1(x)
\end{equation*}
This evolution is driven by the advection equation:
\begin{equation*}
 {\alpha(x)} u_x + u_t = 0
\end{equation*}
where $u_t$ and $u_x$ denote the parital derivaties of $u$ with respect to time $t$ and space $x$, respectively. The velocity field $\alpha(x)$ dictates the rate at which the shape of $u$ propagates. In our formulation, we require {$\alpha$} to be a function belonging to a reproducing kernel Hilbert space (RKHS). Using the method of characteristics \citet{book_pde}, this PDE reduces to a system of ordinary differential equations (ODEs).  Let $x(t)$, denote the spatial coordinate along a characteristic curve; for notational simplicity, we will frequently write $x$ in place of $x(t)$. The reduced 
ODE system is given by:
\begin{equation}\label{eq:sysodes}
\begin{array}{ll}
    \frac{\partial x}{\partial t} = \dot{x} = \alpha(x), &
    \frac{\partial u}{\partial t}=\dot{u} =0
\end{array}
\end{equation}
Equation \eqref{eq:sysodes} demonstrates that $u$ is constant along the characteristics. Refer to Figure \ref{fig:mofc} for a description of our characteristics and to explicitly show what we aim to learn.

\subsection{Deriving the Objective Function}

Given a set of $n$ training pairs $\{(\phi_0^{(i)}, \phi_1^{(i)})\}_{i=1}^n$, where we define $\phi_0^{(i)}=\phi_0(x_i(0))$ and $\phi_1^{(i)}=\phi_1(x_i(1))$, we want to formulate an objective function that encourages each transported point to match its target location at time $t=1$. Therefore, we define the following objective function to minimize:
\begin{equation}\label{eq:Obj_func}
    J(\alpha) = \frac{1}{n}\sum_{i=1}^n \left[\phi_1(x_i(1)) - \phi_0(x_i(0)) \right]^2 + \lambda \|\alpha\|^2,
\end{equation}
subject to the constraint $\dot{x}_i(t) = \alpha(x_i(t))$, for $t \in [0,1]$, where $\lambda > 0$ is a regularization parameter. By the Fundamental Theorem of Calculus, the spatial position at $t=1$ is given by:
\begin{equation*}
    x_i(1) = x_i(0) + \int_0^1 \alpha(x_i(t)) \, dt.
\end{equation*}
Because $x_i(1)$ depends implicitly on the velocity field $\alpha$, computing the gradient of $J(\alpha)$ to find the optimal $\alpha$ requires specific care. Sections \ref{sec:RKHS} and \ref{sec:char} detail the explicit derivation of this gradient.

\subsection{Reproducing Kernel Hilbert Spaces}\label{sec:RKHS}
We employ theory of reproducing kernel Hilbert spaces (RKHSs) in order to learn $\alpha$. Let $\mathcal{X}$ be a nonempty set. A Hilbert space of real-valued functions on $\mathcal{X}$ that allows the pointwise evaluation at any point $x\in X$ to be a continuous linear functional in its norm topology is called an RKHS \citet{ai_rkhs}.
\begin{definition}[Positive Definite Kernel]\label{def:pdkernel}
Let $\mathcal{X}$ be a non-empty set. A function $k: \mathcal{X}\times\mathcal{X} \rightarrow\mathbb{R}$ is called a positive definite kernel on $\mathcal{X}$ if and only if it is symmetric, that is $k(x, y) = k(y,x)$ for any $x, y \in \mathcal{X}$, and positive definite, that is
\begin{equation*}
    \sum_{i=1}^n \sum_{j=1}^n c_i c_j k(x_i, x_j) \geq 0
\end{equation*}
for any $n >0$, any choice of $n$ points $x_1, x_2, \dots, x_n \in \mathcal{X}$, and any choice of real numbers $c_1, \dots, c_n \in \mathbb{R}$. 
\end{definition}
\begin{definition}[RKHS]
Let $\mathcal{X}$ be a set and $\mathcal{H} \subset \mathbb{R}^{\mathcal{X}}$ (that is, let $\mathcal{H}$ be a subset of functions that take elements from $\mathcal{X} \rightarrow \mathbb{R}$) where $\mathcal{H}$ is a class of functions that form a Hilbert space with inner product $\langle \cdot, \cdot \rangle_{\mathcal{H}}$. The function $k:\mathcal{X}\times \mathcal{X} \mapsto \mathbb{R}$ is called a reproducing kernel of $\mathcal{H}$ if
\begin{enumerate}
    \item $\mathcal{H}$ contains all functions of the form
    \begin{equation*}
        \begin{array}{ll}
            \forall{\textbf{x}} \in \mathcal{X}, & k_{x}: y \mapsto k(x,y).
        \end{array}
    \end{equation*}
    \item For every $x\in \mathcal{X}$ and $f\in \mathcal{H}$ the reproducing property holds:
    \begin{equation*}
        f(x) = \langle f, k_{x}\rangle_{\mathcal{H}}.
    \end{equation*}
\end{enumerate}

\end{definition}
A fundamental property of an RKHS is that there exists a continuous feature mapping $\Psi:~\mathcal{X}~ \rightarrow~\mathcal{H}$ such that the kernel evaluates the inner product in $\mathcal{H}$:
\begin{equation*}
k(x,x') = \langle \Psi(x), \Psi(x') \rangle_{\mathcal{H}}.
\end{equation*}
While this feature space is often infinite-dimensional, Bochner's Theorem allows us to approximate shift-invariant kernels using finite-dimensional mappings. In this work, we utilize the Gaussian kernel, 
\begin{equation*}
k(x,x') = \exp{\frac{-\|x-x'\|^2} {2\sigma^2}}, 
\end{equation*}
where $\sigma$ is the bandwidth parameter. To approximate this kernel explicitly, we apply random Fourier features \citet{rahimi2007random}. We draw $D$ frequency scalars (where $D$ is the number of features), $\omega_1, \dots, \omega_D$, independently from a standard normal distribution, $\omega \sim \mathcal{N}(0,1)$. By evaluating the trigonometric transformations scaled by our bandwidth $\sigma$, we define our explicit feature map $\gamma(x) \in \mathbb{R}^{2D}$ (where $p=2D$) as the concatenation of cosine and sine functions:
\begin{equation}\label{eq:rff}
    \gamma(x) = \frac{1}{\sqrt{D}} 
    \begin{bmatrix}
        \cos(\omega_1 x / \sigma) \\
        \vdots \\
        \cos(\omega_D x / \sigma) \\
        \sin(\omega_1 x / \sigma) \\
        \vdots \\
        \sin(\omega_D x / \sigma)
    \end{bmatrix}.
\end{equation}
By combining Bochner's Theorem with the strong law of large numbers, the empirical inner product of these random features converges to the exact kernel function as the dimension $D$ approaches infinity. Consequently, for any $x, x' \in \mathcal{X}$, we have:
\begin{equation*}
\begin{array}{ll}
    \lim_{D \rightarrow +\infty} \gamma(x)^T \gamma(x') = k(x,x'), &\text{a.s}.
    \end{array}
\end{equation*}
Hence, we can express our target function $\alpha \in \mathcal{H}$ as a linear combination of these features using learnable weights $\beta \in \mathbb{R}^p$:
\begin{equation*}
\alpha(x) = \beta^T \gamma(x).
\end{equation*}
This reduces the infinite-dimensional representation of $\alpha$ to a low-dimensional, finite parameterization. Consequently, the objective function, Equation~\eqref{eq:Obj_func}, becomes:
\begin{equation}
\label{eq:beta_obj}
    J(\beta) =  \frac{1}{n}\sum_{i=1}^n  \left[ \phi_1(x_i(1))-\phi_0(x_i(0))\right]^2 + \lambda \beta^T\beta,
\end{equation}
with the constraint
\begin{equation*}
\dot{x}_i = \beta^T\gamma(x_i).
\end{equation*}
The vector $\beta$ is then learned directly.

\subsection{Learning the Characteristics}\label{sec:char}
Computing the gradient of the objective function, $\nabla_\beta J$ using Equation \eqref{eq:beta_obj}, presents a computational challenge because the learnable parameters $\beta$ are implicitly nested within the integration of the final state, $x_i(1)$. To overcome this, we frame the gradient computation as an optimal control problem by augmenting the objective function with continuous Lagrange multipliers, $p_i(t)$. By applying integration by parts to the system's dynamical constraint ($\dot{x}_i = \alpha$), we derive a secondary ODE governing the evolution of the multiplier, $\dot{p}_i(t)$, subject to a defined terminal condition, $p_i(1)$. This formulation enables a continuous backpropagation procedure; by solving the resulting adjoint ODE for $p_i(t)$ backwards in time from $t=1$ to $t=0$, we can efficiently compute the exact gradients in a continuous-time framework that mathematically mirrors backpropagation through time in deep learning \citet{chen2018neural}. Consider the functional:
\begin{equation*}  
\psi\bigg(\beta^T\gamma(\cdot),x_0\bigg) = \frac{1}{n}\sum_{i=1}^n \psi_i\bigg(\beta^T \gamma(\cdot),x_0\bigg),
\end{equation*}
with 
\begin{equation*}
    \psi_i\bigg(\beta^T\gamma(\cdot),x_0\bigg) = \Big[ \phi_1(x_i(1))-\phi_0(x_i(0))\Big]^2,
\end{equation*}
for $i = \lbrace 1,\dots, n\rbrace$ under the constraints defined by the $\dot{x}(t)=\alpha(x)= \beta^T\gamma(x(t))$ and $x(0) = x_0$. Then for each $i$:
\begin{equation*}
\begin{array}{ll}
    \dot x_i(t) =\beta^T \gamma(x_i(t)),\\
    x_i(0)  =x_i(0).
\end{array}
\end{equation*}
Next, we augment our objective function with Lagragian methods to replace this constrained optimization problem and define the Lagrange multipliers $p_i(t)$, $t \in [0,1]$ and 
\begin{equation*}
   \begin{array}{ll}
    J_i(\beta) 
    = & \psi_i\bigg(\beta  ^T\gamma(\cdot),x_0\bigg)+\int_0^{1} p_i(t) \bigg(\dot x_i(t) - \beta^T\gamma(x_i(t)) )\bigg)dt, 
    \end{array}
\end{equation*}
and compute the derivative of each term with respect to $\beta$. Refer to the Appendix \ref{App:derive_obj_func} for the completion of the proof.
Our objective function gradient with respect to $\beta$ becomes:
\begin{equation*} \label{eq:grad_beta}
    \nabla_{\beta} J(\beta) =     - \frac{1}{n}\sum_{i=1}^n  \int_0^{1} p_i(t) \gamma (x_i(t)) dt + 2\lambda \beta,
\end{equation*}
where $p_i(t)$ satisfies the adjoint ODE and terminal condition:
\begin{equation*} \label{p_ode_2}
\begin{array}{ll}
     \dot{p_i}(t) = -p_i(t)\beta^T\frac{\partial}{\partial x} \gamma(x_i(t) ), \\
    p_i(1)  = 2[\phi_0(x_i(0) - \phi_1(x_i(1))]\frac{\partial}{\partial x} \phi_1(x_i(1)).
\end{array}
 \end{equation*}
The algorithm is detailed in Algorithm \ref{alg:gd}.

\begin{algorithm}[h!]
\caption{DiffTW using Gradient Descent}
\label{alg:gd}
\begin{algorithmic}[1] 
\State Initialize $\beta$, stopping criteria
\For{iteration = 1 to stopping criteria} 
    \State Integrate $x(t=1)$ using $\beta$
    \State Compute $p(1)$ using $x(1)$
    \State Integrate $p(1)$ backwards in time to obtain $p(0)$
    \State Compute $\nabla J_{\beta}$ using quadrature
    \State $\beta_{\text{new}} = \beta_{\text{old}} - \eta \nabla J_{\beta}$
\EndFor
\State \Return $\beta_{\text{new}}$
\end{algorithmic}
\end{algorithm}

\subsection{Projection of the Kernel Function} \label{Sec:proj_kernel} 

As part of our approach, we require that $\alpha$ vanishes at the boundaries, meaning $\alpha(0) = 0$ and $\alpha(1) = 0$. This mirrors the endpoint constraints used in standard DTW. Additionally, it ensures that the endpoints of the signals do not shift independently, allowing the learned dynamics ($\dot{x} = \alpha(x)= \beta^\top \gamma(x)$) to focus on meaningful differences in the body of the signals. To enforce these boundary conditions, we construct a modified RKHS, $\mathcal{H}_0 \subset \mathcal{H}$, that removes any components in the RKHS that are nonzero at the endpoints ($x=0$ and $x=1$). We do this by projecting out the subspace spanned by the feature vectors evaluated at the boundaries. The resulting kernel ensures that any function drawn from it automatically satisfies the zero-boundary conditions. We define the RKHS $\mathcal{H}_0 \subset \mathcal{H}$ as:
\begin{equation*}
    \begin{array}{ll}
    \mathcal{H}_0 = \lbrace f \in \mathcal{H}, f(0) = 0 \text{ and } f(1) = 0 \rbrace ,\\
    \langle f,g\rangle_{\mathcal{H}_0} = \langle f,g\rangle_{\mathcal{H}}, &g\in \mathcal{H}_0.
    \end{array}
\end{equation*}
By the reproducing property,
\begin{equation*}
    \begin{array}{ll}
         f(0) = 0 \text{ and } f(1) = 0  \\
         \iff \\
         \langle f, k(\cdot, 0)\rangle = 0 \text { and } \langle f, k(\cdot, 1)\rangle = 0.
    \end{array}
\end{equation*}
This implies that $\mathcal{H}_0 = \text{span}\lbrace k(\cdot, 0), k(\cdot, 1) \rbrace^{\perp}$ is the orthogonal complement in $\mathcal{H}$. Let $k$ be a positive definite kernel on $\mathcal{X}$ with a finite-dimensional feature map $\gamma(x) \in \mathbb{R}^p$ from Equation \eqref{eq:rff}, and define $\Gamma = [\gamma(0), \gamma(1)] \in \mathbb{R}^{p \times 2}$. Then the projected kernel is:
\begin{equation*}
    k_0(x,y) = \gamma(x)^\top (I - \Gamma G^{-1} \Gamma^\top) \gamma(y), \quad G = \Gamma^\top \Gamma.
\end{equation*}
Here, $A = I - \Gamma G^{-1} \Gamma^\top$ is the $p\times p$ orthogonal projection matrix onto the subspace of functions vanishing at the boundaries. To see why functions in this space satisfy the boundary constraints, we can verify that $A$ completely annihilates the boundary features:
\begin{equation*}
\begin{array}{ll}
A\Gamma = \Gamma - (\Gamma G^{-1}\Gamma^T)\Gamma = \Gamma - \Gamma = 0,
\end{array}
\end{equation*}
where $0\in p\times 2$. Since $A\gamma(0) = 0$ and $A\gamma(1) = 0$, it follows that $k_0(0,y) = k_0(1,y) = 0$ for all $y \in \mathcal{X}$, and  every $\alpha \in \mathcal{H}_0$ satisfies $\alpha(0) = \alpha(1) = 0$. 
We can now represent the modified kernel as:
\begin{equation*}
    k_0(x,y) = \gamma(x)^\top A\gamma(y).
\end{equation*}
Using spectral decomposition, we can write $A = U\Lambda U^\top$. Here, $U^\top U = I$, $U = [u_1, \dots, u_{p-2}]$ is the matrix of eigenvectors of $A$, and $\Lambda$ is the diagonal matrix of the $p-2$ positive eigenvalues of $A$. We can then define the new feature vector for $k_0$ as:
\begin{equation}\label{eq:feat_map}
    \gamma_0(x) = \Lambda^{1/2} U^\top \gamma(x).
\end{equation}
This new feature vector has a dimension of $p-2$. Since the projection removes two linearly independent constraints, the feature space of the resulting kernel drops by two dimensions. For details on the derivation, the reproducing property, and the matrix form, see Appendix \ref{App:proj_kernel_derivation}.

\section{Experiments}\label{Sec:Exp}

We began by evaluating our algorithm on a synthetic toy problem. After confirming its ability to recover known dynamics, we extended the framework to real-world datasets, which we explore later in this section.

\subsection{Toy Problem} \label{sec:toy}
\begin{figure}[htbp]
    \centering
    \includegraphics[width=0.9\linewidth]{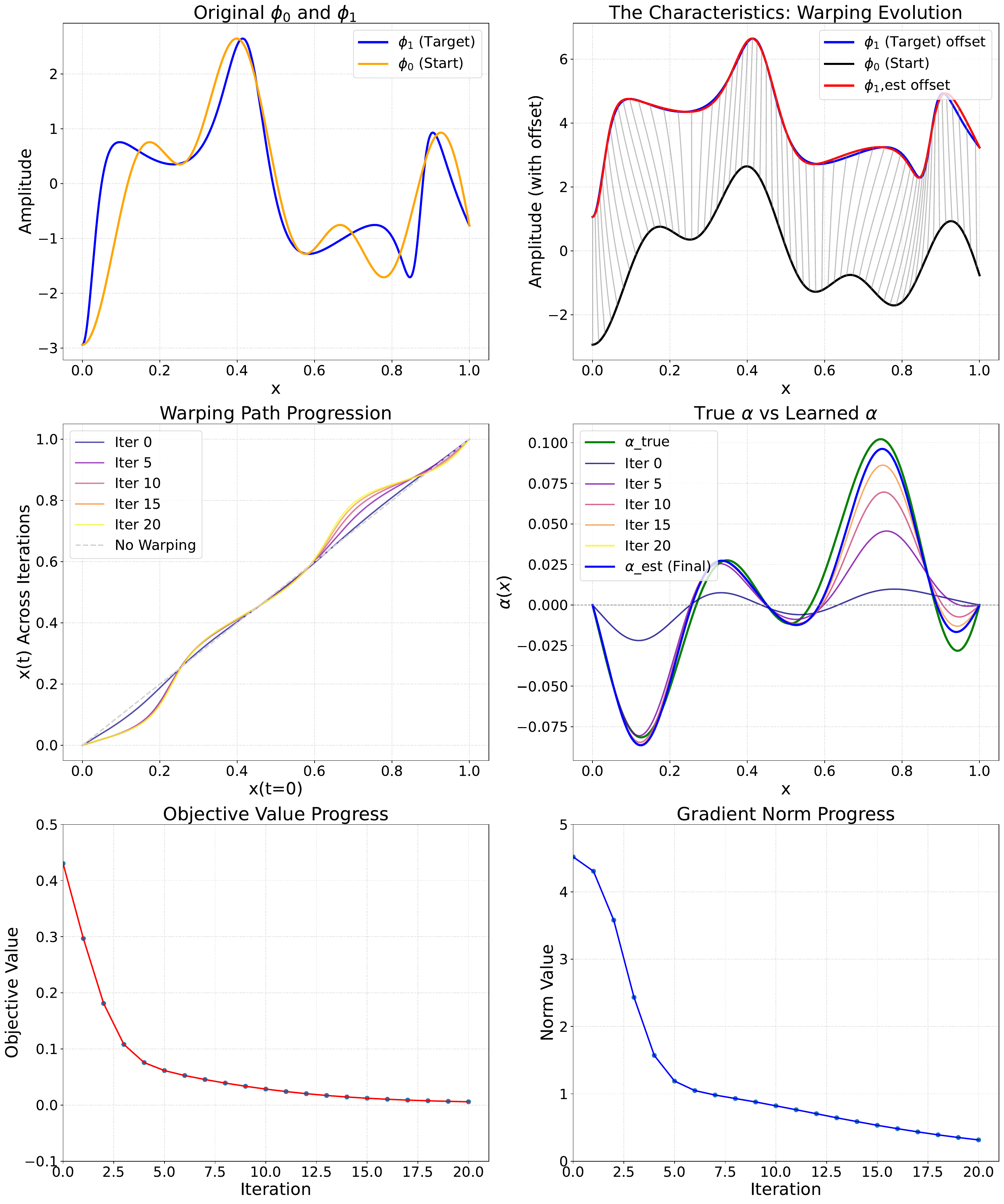}
    
    \caption{Top Left: The original function $\phi_0$ and the target $\phi_1$. Top Right: The warping evolution and characteristic curves, illustrating how individual time points migrate from the source $\phi_0$ to the offset estimated target $\phi_{1,\text{est}}$ across the integration steps. Middle Left: The warping path progression mapping the trajectory dynamics of $x(t)$ across optimization steps, shown for every 5th iteration against the gray dashed identity line. Middle Right: Evolution of the learned velocity field $\alpha_{\text{est}}$, plotted every 5th iteration against a zero reference line. Bottom Left: Objective value across iterations, demonstrating smooth convergence. Bottom Right: Norm of the gradient decreasing during training, indicating optimization stability.}
    \label{fig:alpha_phi0_phi1}
\end{figure}

To test the algorithm, we first constructed a toy problem using random Fourier features. Let the spatial domain be $\mathcal{X} = [0,1]$. Using random Fourier features, we defined a feature map $\gamma(x) \in \mathbb{R}^{p}$ per Equation \eqref{eq:rff} and sampled coefficients $\beta_0 \in \mathbb{R}^{p}$ to define our initial signal:
\begin{equation*}\label{eq:phi_0}
    \phi_0(x) = \beta_0^\top \gamma(x).
\end{equation*}
We then introduced a second set of coefficients $\beta_1 \in \mathbb{R}^{p-2}$ to define the ground-truth vector field governing the temporal dynamics ($\dot{x} = \alpha(x)$) and the feature map $\gamma_0(x) \in \mathbb{R}^{p-2}$ from Equation \eqref{eq:feat_map}:
\begin{equation*}\label{eq:alpha}
    \alpha_{\text{true}}(x) = \beta_1^\top \gamma_0(x).
\end{equation*}
To ensure smooth, non-trivial temporal dynamics, both $\beta_0$ and $\beta_1$ were sampled using $p=2D$ independent samples from normal distributions scaled proportionally to the feature dimension, where $D=50$. For simplicity, we utilized forward Euler integration and evolved $\phi_0$ through the vector field $\alpha_{\text{true}}$ to generate the target signal $\phi_1$. We trained our algorithm (DiffTW) to recover the vector field using only the start and end signals. Figure~\ref{fig:alpha_phi0_phi1} visualizes the learned vector field $\alpha_{\text{est}}$ alongside the training metrics, and shows the resulting evolved signal ${\phi_{1,\text{est}}}$. The hyperparameters used for this toy problem are detailed in Table~\ref{tab:toy_prob_params}. A standalone interactive demonstration replicating this exact synthetic experiment is available on GitHub.\footnote{\url{https://github.com/vgeneva/DiffTW/blob/main/demo_toy.py}}
After 21 iterations of gradient descent, the algorithm demonstrated fast convergence and successfully reconstructed the underlying dynamics. Encouraged by these findings, we next evaluate our method on real-world datasets, beginning with electrocardiogram (ECG) signals in Section~\ref{sec:ECG}.

\begin{table}[tbhp]
\centering
\caption{Parameters of the DiffTW Toy Problem. The algorithm was optimized via gradient descent using the listed parameters, where $\eta$ is the learning rate and $\lambda$ is the regularization weight, $D$ number of features, $\sigma$ for Bandwidth, Iter. for number of iterations of gradient decent, and Step for integration step size.}
\label{tab:toy_prob_params}
\vspace{10pt}
\begin{tabular}{l c c c c c c}
\hline
\textbf{Parameter} & \textbf{$\eta$} & \textbf{$\lambda$} &  $D$&  $\sigma$ & Iter. & Step\\
\hline
\textbf{Value} & 0.00675 & $1\times 10^{-4}$ & 50 & 0.1 & 21 & 1/20\\
\hline
\end{tabular}
\end{table}

\subsection{Electrocardiogram Data}\label{sec:ECG}

\begin{figure}[htbp]
  \centering
\includegraphics[width=.4\linewidth]{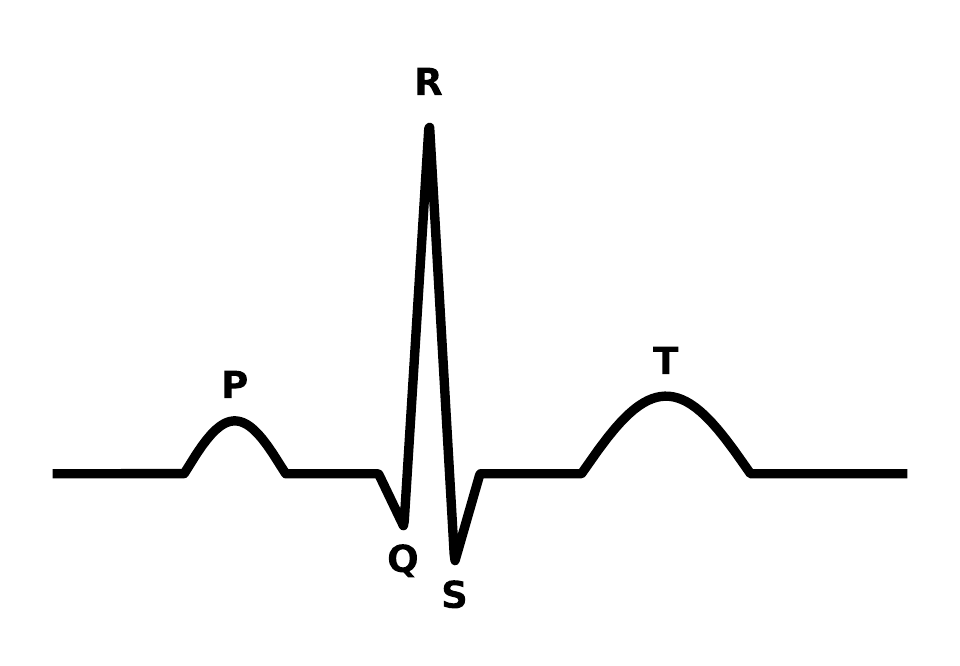}
  \caption{Example of an ECG signal capturing a full cardiac cycle, featuring the P wave, QRS Complex, and T wave.}
  \label{fig:ekg_ex}
\end{figure}

After promising results from Section \ref{sec:toy}, we decided to conduct analysis on electrocardiogram (ECG) signals. ECG records the electrical activity of the heart. As shown in Figure \ref{fig:ekg_ex}, a standard cardiac cycle consists of distinct morphological features: the P wave, the sharp QRS complex, and the T wave. For time series modeling, ECGs present signals that contain a complex mixture of smooth variations and sharp peaks.
To evaluate our algorithm, we used the MIT-BIH Arrhythmia Database \citet{MITBIH}, which is one of the most widely used datasets for ECG research. It contains 48 half-hour recordings of two-channel ECG signals from 47 different patients. The dataset includes a variety of heart arrhythmias and normal heartbeats. 
\subsubsection{An Example: ECG Signal Alignment}
\begin{figure}[htbp]
    \centering
    \includegraphics[width=0.9\linewidth]{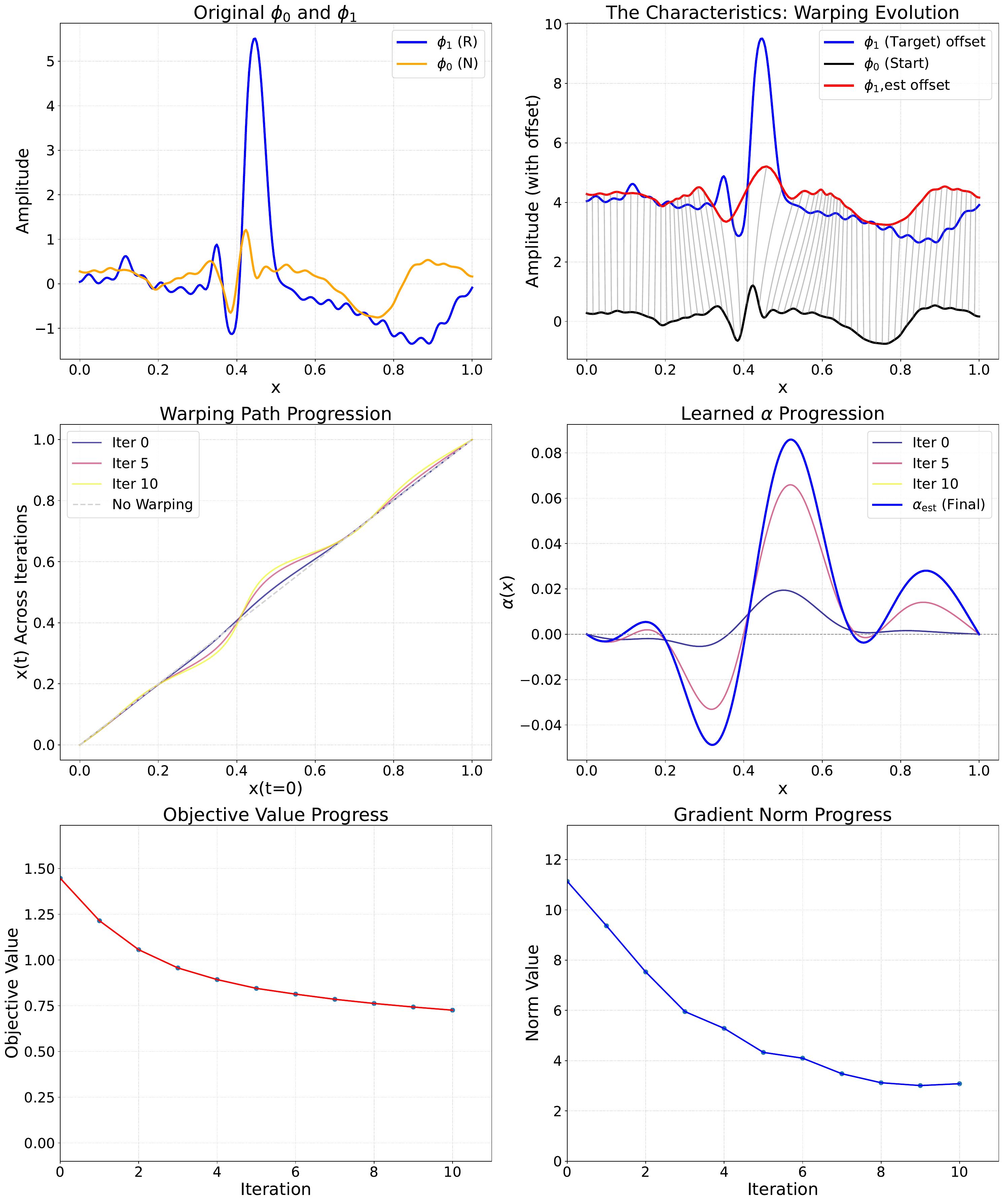}
    \caption{Top Left: The ECG input signals $\phi_0$ (N) and the target $\phi_1$ (R). Top Right: The warping evolution and characteristic curves. Middle Left: Warping path progression mapping the trajectory dynamics of $x(t)$ across 11 optimization iterations. Middle Right: Evolution of the learned velocity field $\alpha_{\text{est}}$ over 11 iterations. Bottom Left: Objective value across iterations, showing decreasing values. Bottom Right: Norm of the gradient decreasing during training. }
    \label{fig:R_N_results}
\end{figure}

Figure \ref{fig:R_N_results} presents two processed  ECG signals (refer to Section \ref{sec:ecg_preprocessing} for processing details), labeled $R$ (right bundle branch block heartbeat) and $N$ (normal heartbeat) along with results of applying our method. 
For convenience, we denote $N$ as $\phi_0$ and $R$ as $\phi_1$. Optimization of $\alpha$ exhibited decreasing objective functions, settling within 11 iterations. We highlight the changes from $\phi_0$ (N) evolving to a featured peak in the $\phi_1$ (R) signal. The hyperparameters used for this ECG problem are detailed in Table \ref{tab:RN_params}. The complete code to reproduce this individual ECG signal alignment visualization is available on GitHub.\footnote{\url{https://github.com/vgeneva/DiffTW/blob/main/demo_ecg.py}}
 \begin{table}[h]
\centering
\caption{Parameters of R and N ECG signals for DiffTW. The model was optimized via gradient descent and used the listed parameters where $\eta$ is the learning rate, $\lambda$ our regularization term, $D$ number of features, $\sigma$ for Bandwidth, Iter. for number of iterations of gradient decent, and Step for integration step size. }
\label{tab:RN_params}
\vspace{12pt}
\begin{tabular}{l c c c c c c}
\hline
\textbf{Parameter} & \textbf{$\eta$} & \textbf{$\lambda$} &  $D$&  $\sigma$ & Iter. & Step\\
\hline
\textbf{Value} & $1/2^9$ & $1\times 10^{-4}$ & 50 & 0.1 & 11 & 1/20\\
\hline
\end{tabular}
\end{table}

\subsubsection{ECG Pre-processing and Classification Procedure}\label{sec:ecg_preprocessing}

To establish a rigorous evaluation framework, we mirror the preprocessing and classification ideas proposed by \citet{cardiac2011}. The authors proposed a DTW-based approach for arrhythmic ECG beat classification and applied a series of preprocessing steps. Following their protocol to ensure morphological consistency across signals, we first removed low-frequency baseline wanderings using a Butterworth high-pass filter at 0.5 Hz, followed by a low-pass filter at 40 Hz to limit the signal bandwidth. Next, R-peaks were identified, based on cardiologist annotations, to locate individual heartbeats. Full cardiac cycles were then extracted by taking 100 samples before and 155 samples after each R-peak, which captures the P-wave, QRS complex, and T-wave. This yielded 256 sample points per beat. Finally, each beat was z-normalized so that the average of the signal was zero, keeping the beats comparable. This was slightly different from the authors’ method, which scaled the amplitudes to the range 0–1.

To evaluate classification performance, we selected the same eight ECG heartbeat classes as in \cite{cardiac2011}. The specific classes, along with the distribution of the training and testing sets, are summarized in Table \ref{tab:ecg_labels}. 
\begin{table}[!h]
\footnotesize
\caption{ECG Labels. Normal (N), Premature ventricular contraction (V), Atrial premature beat (A), Right bundle branch block beat (R), Left bundle branch block beat (L), Ventricular escape beat (E), Junctional (nodal) premature beat (J), and Junctional (Nodal) escape beat (j) for classification.}\label{tab:ecg_labels}
\vspace{12pt}
\begin{tabular*}{\textwidth}{@{\extracolsep{\fill}}lrrrrrrrr@{}}
\hline
Class label & 
\multicolumn{1}{c}{N} & 
\multicolumn{1}{c}{V} & 
\multicolumn{1}{c}{A} & 
\multicolumn{1}{c}{R} & 
\multicolumn{1}{c}{L} & 
\multicolumn{1}{c}{E} & 
\multicolumn{1}{c}{J} & 
\multicolumn{1}{c}{j} \\
\hline
Test Signals & 1000 & 982 & 1000 & 500 & 500 & 95 & 40 & 202 \\
Train Signals & 10 & 10 & 10 & 10 & 10 & 10 & 10 & 10 \\ 
\hline
\end{tabular*}
\end{table}
To establish a benchmark for our DiffTW algorithm, we reproduced the standard DTW baseline results. While the authors in \citet{cardiac2011} utilized a class-wise classification approach and explored sub-sampling to improve detection speed, we opted for a more direct comparison. We performed our analysis without sub-sampling and adopted a standard $k$-nearest neighbor ($k=1$) classification strategy. Specifically, distances are computed between a given test beat and all training beats across the arrhythmia classes, and the test beat is assigned the label of the training beat that yields the minimum distance. To ensure a straightforward comparison, the distance metrics for both frameworks are derived from their respective alignment processes. For traditional unconstrained DTW, the distance is defined as the minimum cumulative alignment cost calculated along an optimal discrete warping path. Let $\phi_0$ and $\phi_1$ represent two discrete ECG signals. Standard DTW calculates the distance by finding an optimal discrete warping path $\pi$ that minimizes the accumulated Euclidean distance between the aligned time points:
\begin{equation}\label{eq:dtw_dist}
    D_{\text{DTW}}(\phi_0, \phi_1) = \min_{\pi} \sum_{k=1}^{K} \|\phi_0(i_k) - \phi_1(j_k)\|^2
\end{equation}
where $K$ is the total length of the warping path, and $(i_k, j_k)$ represents the pair of mapped indices at step $k$. This minimization is solved via dynamic programming, subject to standard boundary, monotonicity, and step-size continuity constraints. To mitigate pathological alignments, where a single point in one signal is mapped to a large section of the other, constrained DTW is also used as a baseline. Constrained DTW modifies the optimization problem by introducing a global warping window $w$ (such as a Sakoe-Chiba band). This imposes an additional restriction on the allowable paths $\pi$, such that $\vert{}i_k - j_k\vert{} \le w$ for all steps $k$ \cite{dtwcode}.

Conversely, DiffTW defines its similarity measure directly through its continuous optimization framework. The distance corresponds to the minimum value achieved by the objective function $J(\beta)$ from Equation \eqref{eq:Obj_func} at convergence or stopping criteria:
    \begin{equation}\label{eq:difftw_dist}
        D_{\text{DiffTW}}(\phi_0, \phi_1) = \min_{\beta} J(\beta)
    \end{equation}
This setup provides a direct, distance-based framework to evaluate the accuracy of DiffTW against DTW. Applying these distance metrics within our 1-NN framework yielded strong classification performance on the ECG dataset. Specifically, DiffTW outperformed the unconstrained (by default) DTW baseline, achieving an overall accuracy of 98.80\% compared to DTW's 98.12\%. Having established the framework's effectiveness on both synthetic data and specific ECG morphological alignments, we next extended our analysis to broader datasets. To thoroughly assess DiffTW's performance relative to DTW, we evaluate it across datasets that vary significantly in time-series length and structural complexity, grouping our ECG classification task with the University of California Riverside Time Series Classification Archive, refer to Section \ref{sec:UCR}. The standardized hyperparameter tuning and optimization strategy applied across all of these classification tasks is detailed in Section \ref{Sec:tuning}, and the full benchmark classification results for the UCR archive, are presented in Section \ref{Sec:Results}.

\subsection{University of California Riverside Time Series Classification Archive}\label{sec:UCR} 

To benchmark our algorithm at scale, we utilized the University of California Riverside (UCR) Time Series Classification Archive \citet{UCRpaper}, a standard repository containing 128 datasets. Of these, 85 datasets were included in our experimental analysis. We excluded datasets that contained missing values, inconsistent sequence lengths, multivariate data formats, or an insufficient number of training/testing samples required for proper hyperparameter tuning (as detailed in Section~\ref{Sec:tuning}). Of the 85 datasets selected for testing, 70 were provided already z-normalized, 9 required manual z-normalization (annotated in Table~\ref{tab:datasets}), and  6 contained exceptionally long time series (length $> 1000$) or excessively large sample sizes. These were subsampled before applying DiffTW to maintain computational feasibility (annotated in Table~\ref{tab:datasets}). Detailed results and specific processing annotations for all 85 datasets are provided in Table~\ref{tab:datasets}. For additional metadata, including study contexts and dataset origins, we refer readers to \citet{bakeoff}.

\subsection{Hyperparameter tuning and Broader Testing} \label{Sec:tuning} While our initial ECG experiments utilized fixed values for $\lambda$ and bandwidth (Table~\ref{tab:RN_params}), evaluating the broader UCR archive required a systematic tuning and training strategy. We performed a grid search over 100 log-spaced values for the regularization term $\lambda \in [10^{-2}, 10^3]$, and 10 uniformly spaced bandwidth values between $1/30$ and $1/5$. To tune these parameters, we constructed representative validation pairs for each class using only the training set. Specifically, we averaged two training signals to form the initial signal $\phi_0$, and averaged a second pair of training signals to form the target signal $\phi_1$. By averaging pairs of training signals, we generate stable, representative signals that allow generalization rather than overfitting to an individual possibly noisy signal, a technique supported by sequence averaging frameworks from \citet{petitjean2011global}. We then evaluated each $(\lambda, \text{bandwidth})$ pair using an index-based 2-fold cross-validation strategy. The temporal indices of $\phi_0$ and $\phi_1$ were split equally into even and odd sets. We optimized the weights $\beta$ for 20 iterations on one subset and computed the objective value on the held-out subset, then swapped the sets. The hyperparameters that yielded the minimum average objective value across both folds were selected to process the full training and test datasets. Finally, we replaced standard gradient descent with the Adaptive Moment Estimation (Adam) optimizer \citet{Kingma2015Adam} and conducted all the experiments for up to 10 iterations utilizing GPU acceleration. Detailed hardware specifications and execution environments are provided in Appendix \ref{App:Code}. Refer to Table \ref{tab:big_difftw} for summary of parameters.
 \begin{table}[h]
\centering
\caption{Parameters used for training on GPU acceleration where GS stands for grid search, $\eta$ is the learning rate, $\lambda$ our regularization term, $D$ number of features, $\sigma$ for Bandwidth, Iter. for number of iterations of gradient decent, and Step for integration step size.}
\label{tab:big_difftw}
\vspace{12pt}
\begin{tabular}{l c c c c c c}
\hline
\textbf{Parameter} & \textbf{$\eta$} & \textbf{$\lambda$} &  $D$&  $\sigma$ & Iter. & Step\\
\hline
\textbf{Value} & $1/2^9$ & GS & 50 & GS & 10 & 1/20\\
\hline
\end{tabular}
\end{table}

\subsection{Computational Complexity}\label{Sec:Complexity}

Importantly, Algorithm \ref{alg:gd} computes the transport by integrating the ODE from $t=0$ to $t=1$. This approach yields a computational complexity of $\mathcal{O}(n \cdot K_{\text{steps}} \cdot \text{iter})$, where $n$ is the sequence length (i.e., the number of transported points), $K_{\text{steps}}$ is the number of integration steps, and $\text{iter}$ is the number of optimization steps. In contrast, traditional DTW requires explicitly constructing a pairwise alignment cost matrix, resulting in a quadratic complexity of $\mathcal{O}(n^2)$ with respect to sequence length. By reframing sequence alignment as a continuous dynamical system, our method replaces DTW's quadratic spatial dependence with a linear dependence on $n$. While DTW relies on exhaustive discrete path searching to handle non-linear time distortions, DiffTW captures these distortions efficiently through smooth vector fields. This exchange of exact dynamic programming for continuous integration drastically reduces computational overhead, making our method highly scalable for long sequences and large datasets.

\section{Results}\label{Sec:Results}
Table~\ref{tab:datasets} reports the overall classification accuracy of DiffTW against the standard DTW baseline (using the default unconstrained warping condition and constrained conditions) across 85 UCR time series archive datasets. Accuracy for these baselines was collected from~\cite{UCRpaper}. For the constrained DTW results, the optimal warping window was learned by searching the training set for the size that yielded the highest 1-NN accuracy. The window sizes are available at the UCR Time Series Classification Archive website. Accuracy is defined as the number of correctly classified test signals divided by the total number of test signals. For the 1-NN classification scheme, DTW utilizes the accumulated Euclidean distance along the optimal alignment path as discussed in Equation \eqref{eq:dtw_dist}. To ensure a fair comparison, DiffTW classification was based on the optimized objective value between pairs of signals as discussed in Equation \eqref{eq:difftw_dist}. Because our objective function measures the sum of squared differences between $\phi_0(x_i(0))$ and the transformed $\phi_1(x_i(1))$, it serves as an analogous, continuous counterpart to the discrete DTW distance.

\begingroup
\footnotesize
\begin{longtable}{l r r r r r}
\caption{The classification accuracies for DiffTW, constrianed DTW (labeled as C) and unconstrained DTW (labeled as U)  taken from~\cite{UCRpaper} across all 85 datasets are reported below.  
This accuracy is the overall accuracy for all the test signals, i.e. total correctly labeled divided by total signals.  
The bold results represent the higher or tied score.}
\label{tab:datasets} \\

\hline
Datasets & \multicolumn{1}{c}{Length} & \multicolumn{1}{c}{Label} & \multicolumn{1}{c}{DiffTW} & \multicolumn{1}{c}{DTW (C)} & \multicolumn{1}{c}{DTW (U)}\\ 
\hline
\endfirsthead

\multicolumn{6}{c}{{\tablename\ \thetable{} -- Continued from previous page}} \\
\hline
Datasets & \multicolumn{1}{c}{Length} & \multicolumn{1}{c}{Label} & \multicolumn{1}{c}{DiffTW} & \multicolumn{1}{c}{DTW (C)} & \multicolumn{1}{c}{DTW (U)} \\ 
\hline
\endhead

\hline
\multicolumn{6}{r}{{Continued on next page...}} \\
\endfoot

\hline
\endlastfoot

ACSFOne & 1460 & 10 & 0.0900 & {\bfseries 0.6200} & {\bfseries 0.6400} \\
Adiac & 176 & 37 & 0.2046 & {\bfseries 0.6087} & {\bfseries 0.6036} \\
ArrowHead & 251 & 3 & {\bfseries 0.7943} & {\bfseries 0.8000} & 0.7029 \\ 
Beef & 470 & 5 & 0.6000 & {\bfseries 0.6667} & {\bfseries 0.6333} \\
BeetleFly & 512 & 2 & {\bfseries 0.7000} & {\bfseries 0.7000} & {\bfseries 0.7000} \\
BirdChicken & 512 & 2 & {\bfseries 0.8500} & 0.7000 & 0.7500 \\
BME & 128 & 3 & {\bfseries 1.0} & 0.9800 & 0.9000 \\
Car & 577 & 4 & {\bfseries 0.7833} & 0.7667 & 0.7333 \\
Crop & 46 & 24 & {\bfseries 0.7188} & 0.7117 & 0.6652 \\
CBF & 128 & 3 & 0.9700 & {\bfseries 0.9956} & {\bfseries 0.9967} \\
ChlorineConcentration & 166 & 3 & 0.5029 & {\bfseries 0.6500} & {\bfseries 0.6484} \\
CinCECGTorso & 1639 & 4 & {\bfseries 0.8688} & {\bfseries 0.9304} & 0.6507 \\
Coffee & 286 & 2 & {\bfseries 1.0} & {\bfseries 1.0} & {\bfseries 1.0} \\
Computers & 720 & 2 & 0.5320 & {\bfseries 0.6200} & {\bfseries 0.7000} \\
Earthquakes & 512 & 2 & 0.6547 & {\bfseries 0.7266} & {\bfseries 0.7194} \\
ECG & 256 & 8 & {\bfseries 0.9880} & - & 0.9182 \\
ECG200 & 96 & 2 & {\bfseries 0.8600} & {\bfseries 0.8800} & 0.7700 \\
ECG5000 & 140 & 5 & {\bfseries 0.9284} & 0.9251 & 0.9228 \\
ECGFiveDays & 136 & 2 & 0.7666 & {\bfseries 0.7967} & {\bfseries 0.7677} \\
ElectricDevices\footnotemark[1] & 48\footnotemark[3] & 7 & 0.5471 & {\bfseries 0.6194} & {\bfseries 0.6012} \\
EOGHorizontalSignal & 1250 & 12 & 0.4669 & {\bfseries 0.4751} & {\bfseries 0.5028} \\
EOGVerticalSignal & 1250 & 12 & {\bfseries 0.4558} & {\bfseries 0.4751} & 0.4475 \\
Ethanollevel\footnotemark[1] & 875\footnotemark[2] & 4 & {\bfseries 0.2800} & {\bfseries 0.2820} & 0.2760 \\
FaceAll & 560 & 14 & 0.8041 & {\bfseries 0.8083} & {\bfseries 0.8077} \\ 
FacesUCR & 131 & 14 & {\bfseries 0.9093} & {\bfseries 0.9122} & 0.9049 \\
Fish & 463 & 7 & {\bfseries 0.8857} & 0.8457 & 0.8229 \\ 
FordA\footnotemark[1] & 250\footnotemark[3] & 2 & 0.4644 & {\bfseries 0.6909} & {\bfseries 0.5546} \\ 
FordB\footnotemark[1] & 250\footnotemark[3] & 2 & 0.4580 & {\bfseries 0.6074} & {\bfseries 0.6198} \\
FreezerRegularTrain & 301 & 2 & 0.7663 & {\bfseries 0.9070} & {\bfseries 0.8989} \\
FreezerSmallTrain & 301 & 2 & {\bfseries 0.7193} & 0.6758 & {\bfseries 0.7589} \\   
GunPoint & 150 & 2 & {\bfseries 0.9733} & 0.9133 & 0.9067 \\   
GunPointAgeSpan\footnotemark[2] & 150 & 2 & {\bfseries 0.9842} & 0.9652 & 0.9177 \\
GunPointMaleVersusFemale\footnotemark[2] & 150 & 2 & {\bfseries 0.9937} & 0.9747 & {\bfseries 0.9968} \\
GunPointOldVersusYoung\footnotemark[2] & 150 & 2 & {\bfseries 0.9651} & {\bfseries 0.9651} & 0.8381 \\
Ham & 431 & 2 & {\bfseries 0.6095} & 0.6000 & 0.4667 \\   
HandOutlines\footnotemark[1] & 677\footnotemark[3] & 2 & {\bfseries 0.8703} & 0.8622 & {\bfseries 0.8871} \\
Haptics & 1092 & 5 & {\bfseries 0.4448} & 0.4123 & 0.3766 \\
Herring & 512 & 2 & {\bfseries 0.5938} & 0.5312 & 0.5312 \\   
HouseTwenty\footnotemark[2] & 301 & 2 & 0.6555 & {\bfseries 0.9412} & {\bfseries 0.9244} \\
InlineSkate & 1882 & 7 & {\bfseries 0.3964} & 0.3873 & 0.3836 \\
InsectWingbeatSound & 256 & 11 & {\bfseries 0.5606} & {\bfseries 0.5848} & 0.3551 \\ 
InsectEPGRegularTrain\footnotemark[2] & 601 & 3 & 0.7671 & {\bfseries 0.8273} & {\bfseries 0.8715} \\ 
ItalyPowerDemand & 24 & 2 & {\bfseries 0.9631} & 0.9553 & 0.9504 \\ 
LargeKitchenAppliances & 720 & 3 & 0.5653 & {\bfseries 0.7947} & {\bfseries 0.7947} \\ 
Lightning2 & 637 & 2 & 0.7049 & {\bfseries 0.8689} & {\bfseries 0.8689} \\
Lightning7 & 319 & 2 & {\bfseries 0.7123} & {\bfseries 0.7123} & {\bfseries 0.7260} \\ 
Mallat & 1024 & 8 & 0.8827 & {\bfseries 0.9143} & {\bfseries 0.9339} \\ 
Meat & 448 & 3 & 0.5167 & {\bfseries 0.9333} & {\bfseries 0.9333} \\  
MedicalImages & 99 & 10 & {\bfseries 0.7605} & 0.7474 & 0.7368 \\
MixedShapesRegularTrain & 1024 & 5 & {\bfseries 0.8973} & {\bfseries 0.9089} & 0.8416 \\ 
MixedShapesSmallTrain & 1024 & 5 & {\bfseries 0.8243} & {\bfseries 0.8326} & 0.7798 \\  
MoteStrain & 84 & 2 & {\bfseries 0.8850} & 0.8658 & 0.8347 \\   
NonInvasiveFetalECGThorax1\footnotemark[1] & 375\footnotemark[3] & 42 & {\bfseries 0.7903} & {\bfseries 0.8107} & {\bfseries 0.7903} \\ 
NonInvasiveFetalECGThorax2\footnotemark[1] & 375\footnotemark[3] & 42 & {\bfseries 0.8758} & 0.8710 & 0.8646 \\ 
OliveOil & 570 & 4 & 0.0667 & {\bfseries 0.8667} & {\bfseries 0.8333} \\ 
OSULeaf & 427 & 6 & {\bfseries 0.6364} & 0.6116 & 0.5909 \\   
PhalangesOutlinesCorrect & 80 & 3 & {\bfseries 0.7681} & 0.7611 & 0.7284 \\ 
Plane & 144 & 7 & 0.9905 & {\bfseries 1.0} & {\bfseries 1.0} \\ 
PowerCons & 144 & 2 & {\bfseries 0.9778} & 0.9222 & 0.8778 \\
RefrigerationDevices & 720 & 3 & {\bfseries 0.4507} & 0.4400 & {\bfseries 0.4640} \\
Rock\footnotemark[2] & 2844 & 4 & {\bfseries 0.7800} & 0.6000 & {\bfseries 0.8400} \\
ScreenType & 720 & 3 & 0.3733 & {\bfseries 0.4107} & {\bfseries 0.3973} \\  
SemgHandGenderCh2\footnotemark[2] & 1500 & 2 & 0.5950 & {\bfseries 0.8450} & {\bfseries 0.8017} \\ 
SemgHandMovementCh2\footnotemark[2] & 1500 & 6 & 0.2244 & {\bfseries 0.6378} & {\bfseries 0.5844} \\
SemgHandSubjectCh2\footnotemark[2] & 1500 & 5 & 0.2889 & {\bfseries 0.8000} & {\bfseries 0.7267} \\
ShapeletSim & 500 & 2 & 0.5056 & {\bfseries 0.7000} & {\bfseries 0.6500} \\
ShapesAll & 512 & 60 & {\bfseries 0.8183} & 0.8020 & 0.7683 \\
SmallKitchenAppliances & 720 & 3 & 0.3787 & {\bfseries 0.6720} & {\bfseries 0.6427} \\
SmoothSubspace & 15 & 3 & {\bfseries 0.9667} & 0.9467 & 0.8267 \\
SonyAIBORobotSurface1 & 70 & 2 & {\bfseries 0.7787} & 0.6955 & 0.7255 \\
SonyAIBORobotSurface2 & 65 & 2 & {\bfseries 0.8604} & 0.8594 & 0.8311 \\
StarLightCurves\footnotemark[1] & 512\footnotemark[3] & 3 & 0.8475 & {\bfseries 0.9053} & {\bfseries 0.9066} \\
Strawberry & 235 & 2 & 0.8865 & {\bfseries 0.9459} & {\bfseries 0.9405} \\
SwedishLeaf & 128 & 15 & {\bfseries 0.8624} & 0.8464 & 0.7920 \\
SyntheticControl & 60 & 6 & 0.9700 & {\bfseries 0.9833} & {\bfseries 0.9933} \\
ToeSegmentation1 & 277 & 2 & {\bfseries 0.7807} & 0.7500 & 0.7719 \\ 
ToeSegmentation2 & 343 & 2 & {\bfseries 0.8846} & {\bfseries 0.9077} & 0.8385 \\  
Trace & 275 & 4 & {\bfseries 0.9900} & {\bfseries 0.9900} & {\bfseries 1.0} \\
TwoLeadECG & 82 & 2 & 0.8560 & {\bfseries 0.8683} & {\bfseries 0.9043} \\ 
TwoPatterns & 128 & 4 & 0.9982 & {\bfseries 0.9985} & {\bfseries 1.0} \\
UMD & 150 & 3 & {\bfseries 0.9861} & 0.9722 & 0.9931 \\
Wafer & 152 & 2 & 0.8960 & {\bfseries 0.9955} & {\bfseries 0.9799} \\
Wine & 234 & 2 & 0.4815 & {\bfseries 0.6611} & {\bfseries 0.5741} \\ 
Worms & 900 & 5 & 0.4805 & {\bfseries 0.5325} & {\bfseries 0.5844} \\
WormsTwoClass & 900 & 2 & {\bfseries 0.6753} & 0.5844 & 0.6234 \\ 
Yoga & 426 & 2 & {\bfseries 0.8640} & 0.8440 & 0.8363 \\
\end{longtable}
\endgroup

\footnotetext[1]{Data was subsampled for DiffTW due to large series length or excessively large number of training and testing samples.}
\footnotetext[2]{z-normalized for DiffTW. }
\footnotetext[3]{For DiffTW:Dataset ElectricDevices was subsampled from 96 to 48. Datasets FordA and FordB were subsampled from 500 to 250. Datasets NonInvasiveFetalECGThorax1 and NonInvasiveFetalECGThorax2 were sub-sampled from 750 to 375. Dataset StarLightCurves was subsampled from 1024 to 512. Dataset HandOutlines subsampled from 2709 to 677. Dataset Ethanollevel subsampled from 1751 to 875.}

\subsection{Discussion}
We evaluated DiffTW through a comprehensive benchmarking study of 86 time-series datasets (85 from the UCR archive, plus a specialized ECG heartbeat dataset). To ensure a rigorous comparison, we excluded datasets with missing values, inconsistent lengths, or multivariate components. Classification was performed using a $k$-nearest neighbor ($k=1$) approach, utilizing the respective alignment costs of DiffTW, Equation \eqref{eq:difftw_dist}, and standard DTW, Equation \eqref{eq:dtw_dist}, as distance metrics. For DTW, we used the UCR Archive constrained and unconstrained results to compare.

\begin{figure}[htbp]
    \centering
    \includegraphics[width=0.8\linewidth]{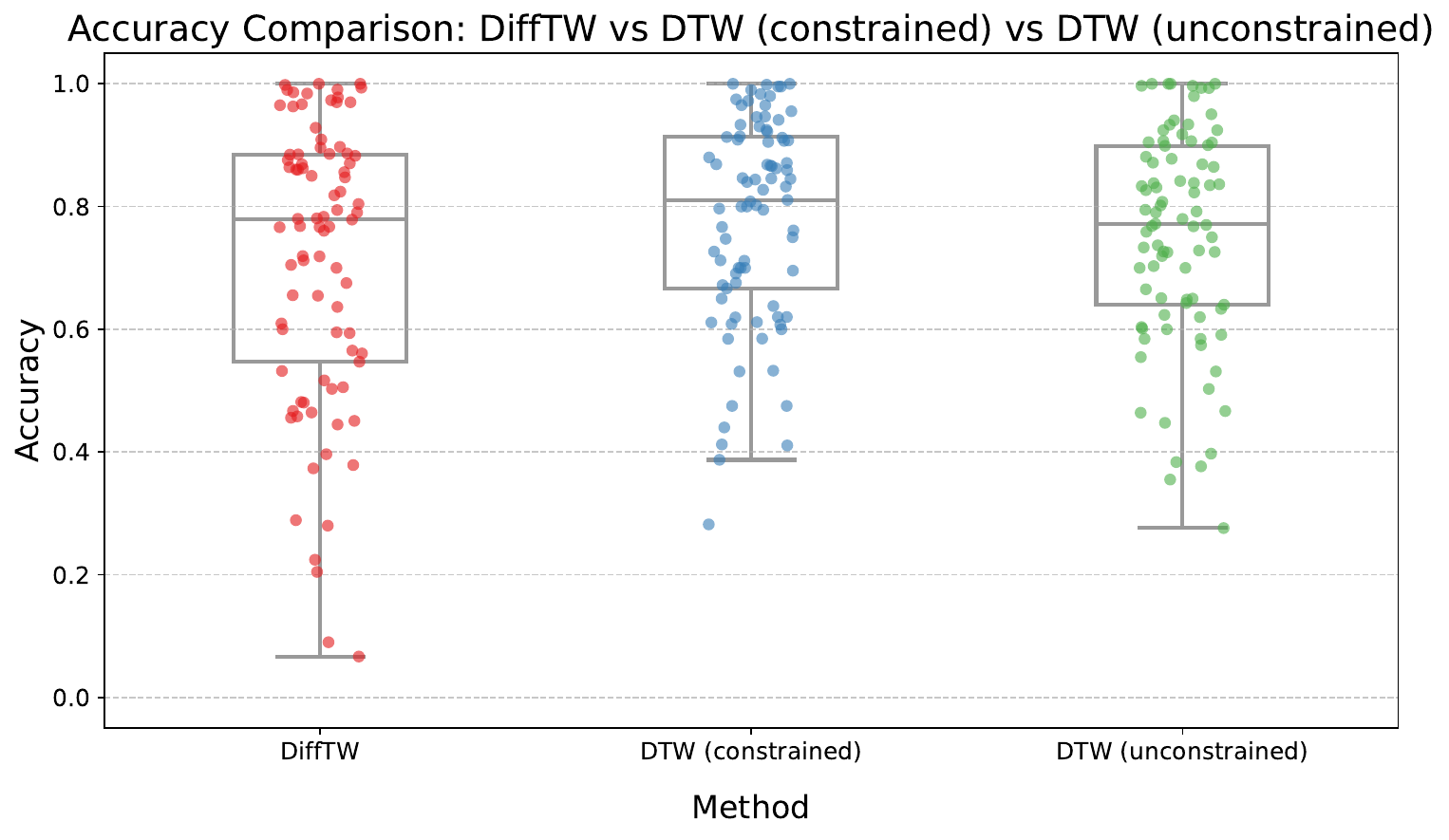}
    \caption{Accuracy comparison across 85 datasets. DiffTW and unconstrained DTW achieve similar performances. Constrained DTW achieve higher accuracy.}
    \label{fig:boxplot}
\end{figure}

In head-to-head accuracy, DiffTW achieved higher scores on 39 datasets, while unconstrained DTW performed better on 43, with 3 ties. DiffTW achieved higher scores on 32 datasets, while constrained DTW performed better on 48 datasets, with 5 ties. A two-sided Wilcoxon signed-rank test conducted in DiffTW and unconstrained DTW which yielded a $p$-value of approximately 0.3214, meaning there is no statistically significant difference between DiffTW and unconstrained DTW. The Wilcoxon test revealed for constrained DTW, a $p$-value of 0.0016 which confirmed that the constrained DTW had statistical significant across this benchmark set. As illustrated in the boxplots in Figure \ref{fig:boxplot}, DiffTW achieves a slightly higher median accuracy of 0.7787 versus unconstrained DTW's median of 0.7719. The Figure also shows the boxplot comparison of DiffTW versus constrained DTW.

\begin{figure}[htbp]
    \centering
    \includegraphics[width=0.7\linewidth]{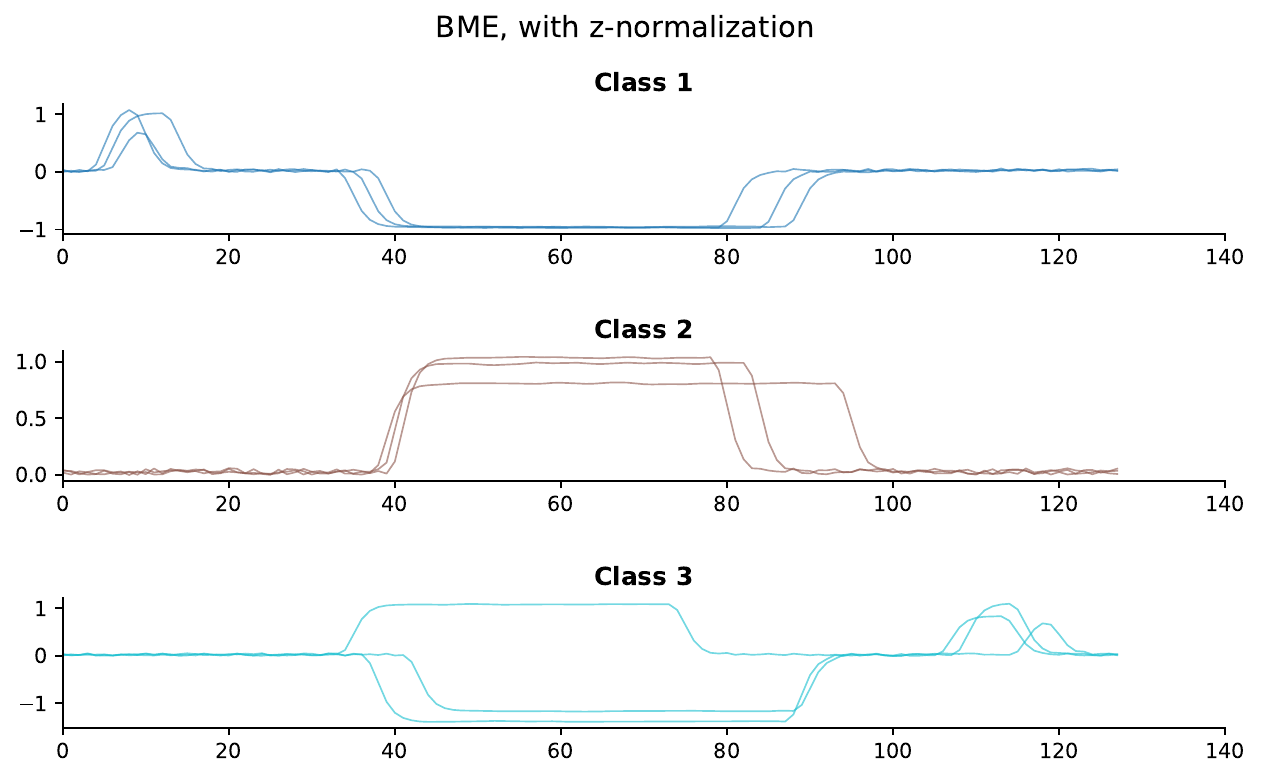}
    \caption{Performance on the BME dataset, highlighting DiffTW's ability to precisely align smooth, localized features like the timing of a specific bell curve.}
    \label{fig:BME}
\end{figure}

The method’s efficacy is particularly evident in tasks requiring precise morphological alignment. In the ECG study, DiffTW successfully aligned key cardiac peaks within 10 iterations by learning an optimal vector field, $\alpha$. This highlights a core strength of the continuous framework: its ability to capture subtle morphological differences, such as those between arrhythmia types, that rigid or purely discrete methods often overlook. This advantage is further demonstrated by the synthetic BME dataset (Figure \ref{fig:BME}). BME consists of three classes characterized by the timing of a small positive bell: one arising at the initial period (Begin), one with no bell (Middle), and one at the final period (End). DiffTW perfectly captured these localized temporal features, achieving 100\% accuracy compared to DTW's constrained value of 98\% and unconstrained of 90\%.

\begin{figure}[htbp]
    \centering
    \includegraphics[width=0.8\linewidth]{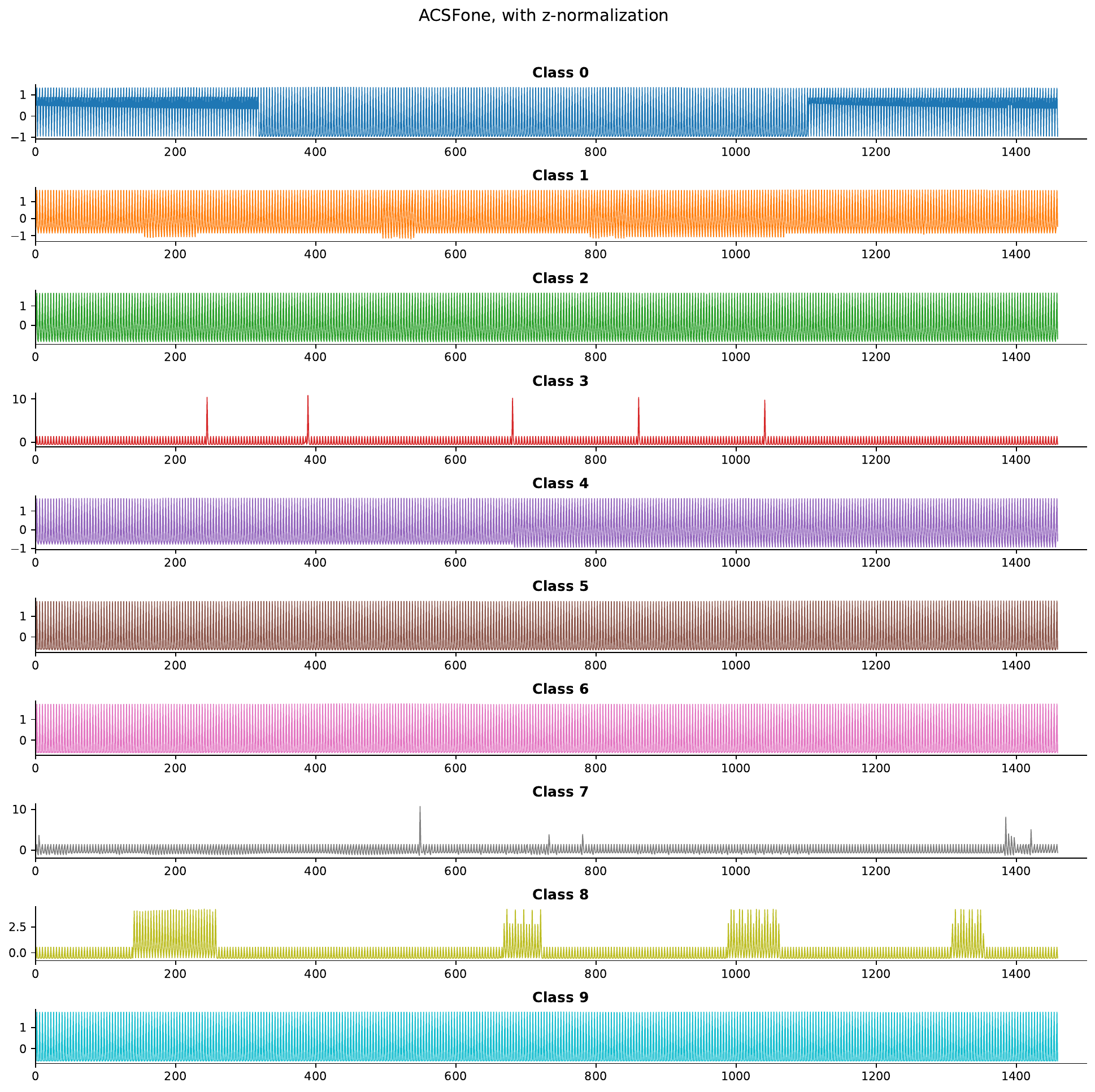}
    \caption{ACSFOne dataset signatures, characterized by sparse high-energy bursts that favor discrete DTW alignment.}
    \label{fig:ACSF1}
\end{figure}

\begin{figure}[htbp]
    \centering
    \includegraphics[width=0.9\linewidth]{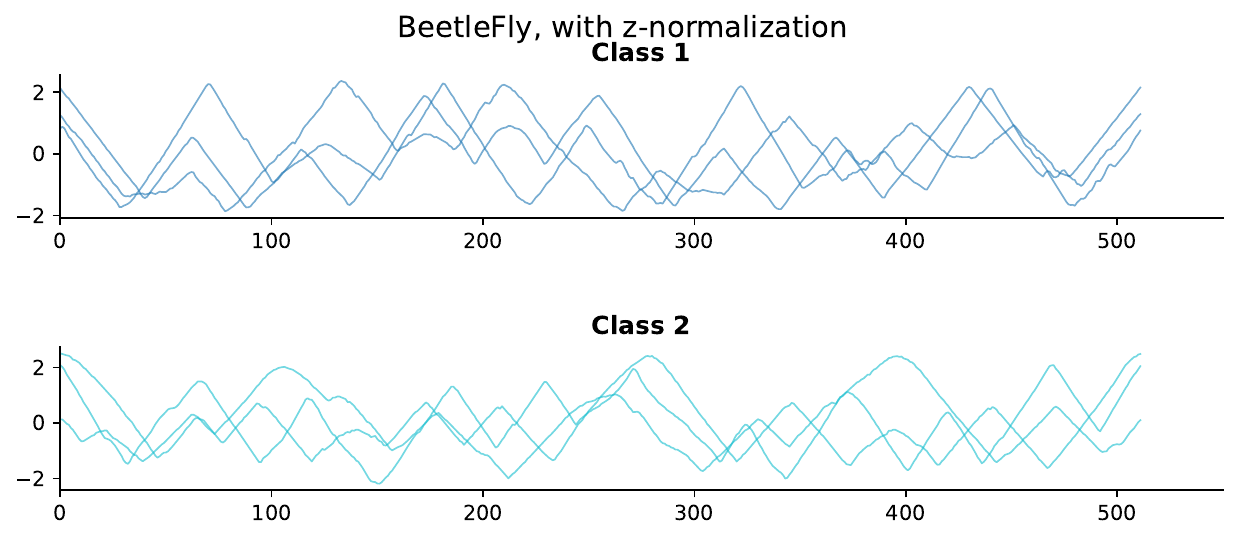}
    \caption{BeetleFly dataset which involves distinguishing an outline of a beetle and a fly, where DiffTW and DTW achieved identical classification accuracy of 70\%.}
    \label{fig:BeeteFly}
\end{figure}

Although DiffTW demonstrates strong comparable performance, our evaluation reveals specific signal characteristics where discrete DTW retains a distinct advantage. A notable example is the ACSFOne dataset (Figure \ref{fig:ACSF1}), where DTW constrained and unconstrained significantly outperformed DiffTW ( {62\% and 64\%, respectively,} vs. 9\% accuracy). ACSFOne signatures are characterized by long idle periods punctuated by sparse, high-energy bursts of power consumption. In such cases, the rigid, step-wise nature of discrete DTW is more robust; the continuous deformation of DiffTW tends to over-smooth or struggle to map isolated, high-magnitude spikes across long stretches of inactivity.
Furthermore, in datasets like BeetleFly (Figure \ref{fig:BeeteFly}), which involve distinguishing between an outline of a beetle and a fly, all methods achieved an equal accuracy of 70\%. This suggests that the primary challenge lies in the complex, overlapping nature of the dataset itself, proving that the moderate accuracy of 70\% is a reflection of the data's inherent difficulty rather than a flaw in the DiffTW framework.

Ultimately, while the datasets where constrained and unconstrained DTW maintained an advantage show that extreme discrete symmetries or sparse bursts still favor traditional warping, the broad success of DiffTW indicates that its continuous vector field framework provides a theoretical alternative for handling complex temporal distortions across diverse domains.

\section{Conclusion}\label{Sec:Conclusion}

In this work, we introduced DiffTW, a continuous temporal alignment framework driven by the integration of smooth vector fields. Extensive benchmarking across 86 datasets demonstrates that DiffTW performs comparably to standard unconstrained discrete DTW providing a mathematically continuous alternative. While highly optimized constrained DTW methods retain a statistically significant accuracy advantage across the broader archive, proves to be particularly effective in capturing and precisely aligning localized morphological features. 

However, our evaluation also identified specific limitations of the continuous approach. In datasets where DTW outperformed DiffTW, the time series were frequently characterized by high-frequency oscillations. These sharp, rapid fluctuations negatively impact the gradient calculations required to smoothly learn the optimal vector field $\alpha$. Additionally, on datasets where the two methods tied or exhibited remarkably similar performance, the underlying classes were often visually indistinguishable. This parity suggests that inherent dataset noise or a lack of clear distinguishing features, rather than the alignment mechanism itself, bounded the maximum achievable accuracy in those instances. 

Future work will explore  hybrid approaches to accelerate convergence, such as utilizing discrete DTW to initialize the continuous optimization. By deriving an initial parameter $\beta$ from the discrete alignment, we hypothesize that this initialization will significantly reduce training time while preserving the benefits of a continuous alignment path. Furthermore, given that DiffTW already achieves linear theoretical complexity with respect to sequence length, subsequent research will focus on empirical runtime analysis and implementation-level optimizations to further accelerate the framework. Finally, further research will investigate what makes one algorithm perform better than the other.

Overall, DiffTW offers a highly flexible alternative to traditional dynamic programming that provides a mathematically grounded, continuous perspective on sequence alignment that performs comparably to standard baselines provided the underlying signals do not contain severe high-frequency noise that disrupts continuous, gradient-based optimization.

\section{Code}%
\label{sec:code}
The computational framework and algorithm proposed in this paper were implemented in \proglang{Python}. To ensure fast and stable continuous trajectory generation, the model architecture, forward Euler integrations, and gradient descent optimization were built utilizing \pkg{tensorflow} and \pkg{tensorflow-probability}. The complete source code and reproducibility instructions are provided in the Appendix \ref{App:Code}.

\section{Funding and Acknowledgments}
This work was supported in part by the National Science Foundation
Research Training Grant (DMS-2136228) and by the National Institute of Health RO1AG021155, R01EY032284, and R01AG027161. 

The authors would like to thank Prof. Eamonn Keogh for helping us identify a bug in our performance evaluation code for DTW. Additionally, we would like to thank all the people who have contributed to the UCR time series classification archive.

\clearpage
\bibliographystyle{jds}
\bibliography{bib}

\clearpage
\appendix

\section{Hardware Environment and Code Execution}\label{App:Code}

To efficiently process the large matrix operations and parallelize the optimization tasks across the 86 evaluated datasets, experiments were executed on a high-performance computing cluster utilizing NVIDIA A40 GPUs (48GB VRAM) running CUDA 12.8. 

All \proglang{Python} scripts, including \texttt{demo\_toy.py}, \texttt{demo\_ecg.py} and \texttt{difftw.py}, as well as the accompanying \texttt{requirements.txt} and datasets, are hosted on our public GitHub repository: \url{https://github.com/vgeneva/DiffTW}. The repo provides example dataframes to use from the UCR archive. To utilize the algorithm on two timeseries, first install the required packages using \pkg{pip}:
\begin{verbatim}
pip install -r requirement.txt
\end{verbatim}

\section{Derivation of the Objective Function}\label{App:derive_obj_func}
\begin{proof}[Proof of the Objective Function]Consider the functional:
\begin{equation*}  
\psi\bigg(\beta^T\gamma(\cdot),x_0\bigg) = \frac{1}{n}\sum_{i=1}^n \psi_i\bigg(\beta^T \gamma(\cdot),x_0\bigg) ,
\end{equation*}
with 
\begin{equation*}
    \psi_i\bigg(\beta^T\gamma(\cdot),x_0\bigg) = \Big[ \phi_1(x_i(1))-\phi_0(x_i(0))\Big]^2 ,
\end{equation*}
for $i = \lbrace 1,\dots, n\rbrace$ under the constraints defined by the $\dot{x}(t)=\alpha(x)= \beta^T\gamma(x(t))$ and $x(0) = x_0$. Then for each $i$:
\begin{equation*}
\begin{array}{ll}
    \dot x_i(t) =\beta^T \gamma(x_i(t)), \\
    x_i(0)  =x_i(0).
\end{array}
\end{equation*}
Next, we augment our objective function with Lagragian methods to replace this constrained optimization problem and define the Lagrange multipliers $p_i(t)$, $t \in [0,1]$ and 
\begin{equation}\label{eq:obj_func_lagrange_mult}
   \begin{array}{ll}
    J_i(\beta) 
    = & \psi_i\bigg(\beta  ^T\gamma(\cdot),x_0\bigg)+\int_0^{1} p_i(t) \bigg(\dot x_i(t) - \beta^T\gamma(x_i(t)) )\bigg)dt ,
    \end{array}
\end{equation}
and compute the derivative of each term with respect to $\beta$.
We begin with the first summand of \ref{eq:obj_func_lagrange_mult}: 
\begin{equation*}
\begin{array}{ll}
     \frac{\partial}{\partial \beta}\psi_i\bigg(\beta ^T\gamma(\cdot),x_0\bigg) 
   & =\frac{\partial}{\partial \beta}\left[\phi_1(x_i(1)) - \phi_0(x_i(0))\right]^2 \\
    & = 2\left[\phi_1(x_i(1)) - \phi_0(x_i(0))\right] \quad \\
    & \cdot \left[ \frac{\partial}{\partial x} \phi_1(x_i(1))\frac{\partial}{\partial \beta} x_i(1) - \frac{\partial}{\partial x} \phi_0(x_i(0)) \frac{\partial }{\partial \beta} x_i(0)  \right] \\
    & = 2\left[\phi_1(x_i(1)) - \phi_0(x_i(0))\right] \left[ \frac{\partial}{\partial x} \phi_1(x_i(1))\frac{\partial}{\partial \beta} x_i(1)  \right] ,
\end{array}
\end{equation*}
using the fact that $x_i(0)$ are constant, thus $\frac{\partial}{\partial \beta} x_i(0)=0$ for each $i$. Next, 
\begin{equation*}
\begin{array}{ll}
    \frac{\partial}{\partial \beta} \int_0^{1} p_i(t) \dot x_i(t) dt 
    &= \frac{\partial}{\partial \beta}\left[p_i(t) x_i(t)\bigg|_0^{1} - \int_0^{1}  \dot{p_i}(t)  x_i(t) dt \right]\\
    & = \frac{\partial}{\partial \beta}\left[p_i(1)  x_i(1)- p_i(0)  x_i(0) - \int_0^{1}  \dot{p_i}(t)x_i(t)  dt\right] \\
    & = p_i(1)  \frac{\partial}{\partial \beta}x_i(1)- p_i(0)  \frac{\partial}{\partial \beta }x_i(0) - \int_0^{1}  \dot{p_i}(t)\frac{\partial}{\partial \beta}x_i(t) dt\\
     & = p_i(1) \frac{\partial}{\partial \beta} x_i(1) - \int_0^{1} \dot{p_i}(t)\frac{\partial}{\partial \beta} x_i(t)dt ,
\end{array}
\end{equation*}
where again we used $\frac{\partial}{\partial\beta} x_i(0) = 0$ and integration by parts in the second line. Lastly, 
\begin{equation*}
\begin{array}{ll}
    -\frac{\partial}{\partial \beta}\int_0^{1} p_i(t)  \beta^T\gamma(x_i(t))dt 
    & =    -\int_0^{1} p_i(t) \left( \frac{\partial}{\partial \beta} \left(\beta^T\gamma(x_i(t))\right) \right) dt \\
    & =    -\int_0^{1} p_i(t) \left(\gamma(x_i(t)) + \beta^T \frac{\partial}{\partial x} \gamma(x_i(t))\frac{\partial}{\partial \beta}x_i(t)\right) dt \notag \\ 
    & =    -\int_0^{1} p_i(t)    \gamma(x_i(t))dt  -\int_0^{1} p_i(t) \beta^T \frac{\partial}{\partial x} \gamma(x_i(t))\frac{\partial}{\partial \beta}x_i(t)dt .
\end{array}
\end{equation*}
Finally, we address the regulation term:
\begin{equation*}
\begin{array}{ll}
      \lambda \frac{\partial}{\partial \beta}  \Vert \beta   \Vert^2   & = \lambda(\frac{\partial}{\partial \beta}\langle \beta, \beta  \rangle)  =\lambda(\langle \frac{\partial}{\partial \beta}\beta, \beta  \rangle+ \langle \frac{\partial}{\partial \beta}\beta, \beta  \rangle) = \lambda (\langle \mathbb{I}, \beta  \rangle+ \langle \mathbb{I}, \beta  \rangle) = 2\lambda \beta .
\end{array}
\end{equation*}
Combining the results above, we obtain:
\begin{equation*}
\begin{array}{ll}
    \frac{\partial}{\partial \beta}J_i(\beta) 
    & = 2\left[\phi_1(x_i(1)) - \phi_0(x_i(0))\right]\left[ \frac{\partial}{\partial x} \phi_1(x_i(1))\frac{\partial}{\partial \beta} x_i(1)  \right]  \\
    &\quad  + p_i(1) \frac{\partial}{\partial \beta} x_i(1)  - \int_0^{1} \dot{p_i}(t)\frac{\partial}{\partial \beta} x_i(t)dt  \\
    &\quad -\int_0^{1} p_i(t)    \gamma(x_i(t))dt \\
    &\quad -\int_0^{1} p_i(t) \beta^T \frac{\partial}{\partial x} \gamma(x_i(t))\frac{\partial}{\partial \beta}x_i(t)dt .
\end{array}
\end{equation*}
We choose:
\begin{equation*}
     \dot{p_i}(t) = -p_i(t)\beta^T \frac{\partial}{\partial x} \gamma (x_i(t )).
\end{equation*}    
then we cancel out two terms and obtain:
\begin{equation*}
\begin{array}{ll}
    \frac{\partial}{\partial \beta}J_i(\beta) 
    &\quad = 2\left[\phi_1(x_i(1)) - \phi_0(x_i(0))\right] \\
    &\quad \cdot \left[ \frac{\partial}{\partial x} \phi_1(x_i(1))\frac{\partial}{\partial \beta} x_i(1)  \right]  \\
    &\quad + p_i(1) \frac{\partial}{\partial \beta} x_i(1)  -\int_0^{1} p_i(t)    \gamma(x_i(t))dt . 
\end{array}
\end{equation*}
Let 
\begin{equation*}
    p_i(1)  = 2\left(\phi_0(x_i(0)) - \phi_1(x_i(1))\right)\frac{\partial}{\partial x} \phi_1(x_i(1)),
\end{equation*}
then our derivative is represented as
\begin{equation*}
    \frac{\partial}{\partial \beta}J_i(\beta) =   -\int_0^{1} p_i(t)    \gamma(x_i(t))dt .  
\end{equation*}

\end{proof}

\section{Projection Kernel Derivation}\label{App:proj_kernel_derivation}
Let $\mathcal{X} \neq \emptyset$ and $k$ be a positive definite kernel function on $\mathcal{X}$ and $H$ be the RKHS of $k$. We will construct an RKHS $H_0 \subset H$ such that
\begin{equation*}
    \begin{array}{ll}
    H_0 = \lbrace f \in H, f(0) = 0 \text{ and } f(1) = 0 \rbrace ,\\
    \langle f,g\rangle_{H_0} = \langle f,g\rangle_H.
    \end{array}
\end{equation*}
Note that since $f\in H$, we can leverage the reproducing property that shows:
\begin{equation*}
    \begin{array}{ll}
         f(0) = 0 \text{ and } f(1) = 0  \\
         \iff \\
         \langle f, k(\cdot, 0)\rangle = 0 \text { and } \langle f, k(\cdot, 1)\rangle = 0.
    \end{array}
\end{equation*}
This implies that $H_0$ is the orthogonal complement in $H$ of $V = \text{span}\lbrace k(\cdot, 0), k(\cdot, 1) \rbrace$. Then for $y\in \mathcal{X}$ the orthogonal projection $\pi_V k(\cdot, y)$ of the function $k(\cdot, y)$ on $V$ is characterized by
\begin{equation}\label{eq:proj}
    \begin{array}{ll}
            \pi_V k(\cdot, y) \in V,\\
            \big\langle k(\cdot, y) - \pi_V k(\cdot,y), k(\cdot, u)\big\rangle = 0 & \text{for }u = 0,1.
    \end{array}
\end{equation}
We can represent the projection of $\pi_Vk(\cdot,y)$ as a linearly combination of the kernel functions of $V$ (since they represent a basis for $V$):
\begin{equation*}
    \pi_Vk(\cdot, y)  = \sum_{v=0}^1 \mu_v(y) k(\cdot, v) = \overline{\mu}^T\overline{k}(\cdot),
\end{equation*}
where $\overline{k}(\cdot) = (k(\cdot, 0), k(\cdot, 1))^T$, $\mu_v(y) \in \mathbb{R}$ and $\overline{\mu} = (\mu_0(y), \mu_1(y))^T$. Substituting $\pi_V k(\cdot, y)$ into Equation \eqref{eq:proj} and using the reproducing property:
\begin{equation*}
    \begin{array}{ll}
       0  & = \big\langle k(\cdot, y) - \pi_V k(\cdot,y), k(\cdot, u)\big\rangle  \\
         & = \big\langle k(\cdot, y) - \sum_{v=0}^1 \mu_v k(\cdot, v), k(\cdot, u)\big\rangle \\
         & = \big\langle k(\cdot, y),k(\cdot, u)\big\rangle - \sum_{v=0}^1  \mu_v\big\langle k(\cdot, v),k(\cdot, u) \big\rangle \\
         & = k(u,y) - \sum_{v=0}^1 \mu_v k(u,v)\\
         \Rightarrow & k(u,y) = \sum_{v=0}^1  \mu_v k(u,v),
    \end{array}
\end{equation*}
for $u=0,1$. We can represent $k(u,y)$ in matrix form for each $u = 0,1$. Let $\overline{k}(y) = (k(0,y), k(1,y))^T$, $\overline{\mu} = (\mu_0(y), \mu_1(y))^T$, then we can represent $\overline{k}(y)$ as
\begin{equation*}
    \begin{array}{ll}
    \overline{k}(y) = G\overline{\mu} \text{ with } G = \begin{bmatrix}
        k(0,0) & k(0,1) \\
        k(1,0) & k(1,1)
    \end{bmatrix}.
    \end{array}
\end{equation*}
Then if we assume $G$ is invertible we obtain
\begin{equation*}
    \begin{array}{ll}
    \overline{\mu} = G^{-1}\overline{k}(y)\\
    \text{and} \\
    \pi_V k(\cdot, y) = \overline{\mu}^T \overline{k}(\cdot) = \overline{k}(y)^T G^{-1} \overline{k}(\cdot), 
    \end{array}
\end{equation*}
since $(G^{-1})^T=(G^T)^{-1} = G^{-1}$. And since 
\begin{equation*}
\begin{array}{ll}
    k(\cdot, y) - \pi_V k(\cdot,y) \in H_0 \\
    \Rightarrow \pi_{H_0} k(\cdot, y) = k(\cdot, y) - \pi_V k(\cdot,y).
    \end{array}
\end{equation*}
We can define the kernel $k_0$ as
\begin{equation}\label{eq:k_0}
    \begin{array}{ll}
    k_0(x,y) & = \langle \pi_{H_0} k(\cdot, y), \pi_{H_0} k(\cdot, x)\rangle \\
    & = \langle k(\cdot, y) - \pi_V k(\cdot,y), k(\cdot, x) - \pi_V k(\cdot,x) \rangle\\
    & = k(y,x) - \langle k(\cdot,y), \pi_V k(\cdot,x) \rangle\\ 
    & - \langle \pi_V k(\cdot,y), k(\cdot,x) \rangle 
     + \langle  \pi_V k(\cdot,y), \pi_V k(\cdot,x) \rangle\\
     & = k(y,x) - \langle k(\cdot, y), \overline{k}(x)^T G^{-1} \overline{k}(\cdot) \rangle\\
     & - \langle \overline{k}(y)^T G^{-1} \overline{k}(\cdot) , k(\cdot, x) \rangle \\
     &+ \langle \overline{k}(y)^T G^{-1} \overline{k}(\cdot) ,\overline{k}(x)^T G^{-1} \overline{k}(\cdot) \rangle.
    \end{array}
\end{equation}
Let $\overline{k}(x)^TG^{-1} = \begin{bmatrix} (\alpha_1(x)& \alpha_2(x) \end{bmatrix} =\overline{\alpha}^T(x)   \in \mathbb{R}^{1\times 2}$, then
\begin{equation*}
    \begin{array}{ll}
         \langle k(\cdot, y), \overline{k}(x)^TG^{-1}\overline{k}(\cdot)\rangle  
        &  =\langle k(\cdot, y),\overline{\alpha}^T(x)\overline{k}(\cdot)\rangle \\
        & = \langle k(\cdot, y), \overline{\alpha}^T(x) (k(\cdot, 0), k(\cdot, 1))^T\rangle \\
        &  = \langle k(\cdot, y), \alpha_1(x) k(\cdot, 0) + \alpha_2(x) k(\cdot, 1)\rangle  \\
        & =\langle k(\cdot, y), \alpha_1 (x) k(\cdot, 0) \rangle + \langle k(\cdot, y) , \alpha_2 (x) k(\cdot, 1)\rangle \\
        & = \alpha_1 (x)k(0,y) + \alpha_2 (x)k(1,y)\\
        & =\overline{k}(y)^T \overline{\alpha}(x) \\
        & = \overline{k}(y)^T (\overline{k}(x)^T G^{-1})^T\\
        & = \overline{k}(y)^T G^{-1} \overline{k}(x).
    \end{array}
\end{equation*}
Similarly, let $\overline{k} (y)^TG^{-1}= \begin{bmatrix} \beta_1(y) & \beta_2(y) \end{bmatrix} =\overline{\beta}^T (y) $:
\begin{equation*}
    \begin{array}{ll}
           \langle \overline{k}(y)^T G^{-1} \overline{k}(\cdot) , k(\cdot, x) \rangle  
           & = \overline{k}(x)^T G^{-1} \overline{k}(y)\\
          & = \overline{k}(y)^T G^{-1} \overline{k}(x),
    \end{array}
\end{equation*}
where the last equality is a property of the inner product. Then finally, 
\begin{equation*}
\small
    \begin{array}{ll}
        \langle \overline{k}(y)^T G^{-1} \overline{k}(\cdot) ,\overline{k}(x)^T G^{-1} \overline{k}(\cdot) \rangle  
         & = \langle \overline{\beta}^T (y)\overline{k}(\cdot), \overline{\alpha}^T(x) \overline{k}(\cdot) \rangle \\
         & = \langle \beta_1(y) k(\cdot, 0) + \beta_2 (y) k(\cdot, 1) , \alpha_1 (x)k(\cdot, 0) + \alpha_2 (x)k(\cdot, 1) \rangle\\
         & = \langle \beta_1(y) k(\cdot, 0), \alpha_1(x) k(\cdot,0)\rangle + \langle \beta_1(y)k(\cdot, 0), \alpha_2 (x) k(\cdot,1)\rangle\\
         &+\langle \beta_2(y) k(\cdot, 0), \alpha_1 k(\cdot,1)\rangle+\langle \beta_2(y) k(\cdot, 0), \alpha_2 (x) k(\cdot,1)\rangle\\
         & = \beta_1(y)\alpha_1 (x) k(0,0) + \beta_1(y)\alpha_2(x) k(0,1) \\
         &+ \beta_2(y) \alpha_1(x) k(1,0) + \beta_2(y) \alpha_2k(x) (1,1) \\
         &=\begin{bmatrix}
             \beta_1 & \beta_2
         \end{bmatrix} \begin{bmatrix}
             k(0,0) & k(1,0)\\
             k(1,0) & k(1,1) 
         \end{bmatrix} \begin{bmatrix}
             \alpha_1 & \alpha_2
         \end{bmatrix}^T\\
         & = \overline{\beta}^T G \overline{\alpha}\\
         & = \overline{k}(y)^TG^{-1} G (\overline{k}(x)^TG^{-1})^T\\
         & = \overline{k}(y)^TG^{-1} G G^{-1}\overline{k}(x)\\
         & = \overline{k}(y)^TG^{-1}\overline{k}(x).
    \end{array}
\end{equation*}
So Equation~\eqref{eq:k_0} becomes: 
\begin{equation}
\begin{array}{ll}\label{eq:new_kernel}
     k_0(x,y) & = k(x,y) - \overline{k}(y)^T G^{-1} \overline{k}(x)  \\
     & =k(x,y) - \overline{k}(x)^T G^{-1} \overline{k}(y).
\end{array}
\end{equation}
\begin{proposition}{The pd kernel of $H_0$ is $k_0$}
\begin{proof}
    We want to show that $k_0 \in H_0$ and the reproducing property holds. From Equation~\eqref{eq:new_kernel}, we obtain the function
    \begin{equation*}
        k_0(\cdot,y) = k(\cdot,y) - \overline{k}(\cdot)^T G^{-1} \overline{k}(y)
    \end{equation*}
    which is a linear combination of $k(\cdot,y)$, $k(\cdot, 0)$ and $k(\cdot, 1)$, all of which belong to $H$, our RKHS. Moreover:
\begin{equation*} \begin{array}{ll} \begin{bmatrix} k_0 (0,y) \\ k_0 (1,y) \end{bmatrix} 
&= \begin{bmatrix}
            k(0,y)\\
            k(1,y) 
        \end{bmatrix}-
        \begin{bmatrix}
            k(0,0) & k(0,1) \\
            k(1,0) & k(1,1) 
        \end{bmatrix} G^{-1} 
        \begin{bmatrix}
            k(0,y)\\
            k(1,y)
        \end{bmatrix} = 0,
        \end{array}
    \end{equation*}
    since $GG^{-1} = I$. Therefore, $k_0(\cdot,y) \in H_0$. Next, we verify the reproducing property holds. Take any $f\in H_0$
    \begin{equation*}
    \small
    \begin{array}{ll}
      \langle f, k_0(\cdot, y) \rangle_H 
      & = \langle f, k(\cdot,y) - \overline{k}(\cdot)^T G^{-1} \overline{k}(y)\rangle_H  \\
         & = \langle f, k(\cdot,y) \rangle - \langle f, \overline{k}(\cdot)^T G^{-1} \overline{k}(y)\rangle_H \\
         & = \langle f, k(\cdot,y) \rangle - \langle f, \overline{k}(y)^T G^{-1} \overline{k}(\cdot)\rangle_H \\
         & = f(y) - \langle f, \overline{\beta}^T \overline{k}(\cdot) \rangle_H\\
         & = f(y) - \langle f, \beta_1 (y)k(\cdot, 0) -\beta_2(y) k(\cdot,1) \rangle_H \\
         & = f(y) - \beta_1(y)\langle f, k(\cdot, 0)\rangle_H -\beta_2(y) \langle f, k(\cdot, 1) \rangle_H\\
         & = f(y) - \beta_1(y) f(0) -\beta_2(y) f(1) \\
         & = f(y) -\overline{\beta}^T \begin{bmatrix}
             f(0) \\
             f(1)
         \end{bmatrix}\\
         & = f(y) - \overline{k}(y)^TG^{-1} \begin{bmatrix}
             f(0) \\
             f(1)
         \end{bmatrix}\\
         & = f(y),
    \end{array}
    \end{equation*}
since $f\in H_0$, then $f(0) = f(1) =0$. Therefore, $\langle \cdot, \cdot \rangle \in H$ is the inner product in $H_0$. 
\end{proof}
\end{proposition}
We observe for $\lbrace x_1, x_2, \dots, x_n \rbrace$, for $x_i \in \mathcal{X}$, then 
\begin{equation*}\begin{array}{ll}
k_0(x_i,x_j) & = k(x_i,x_j) - \overline{k}(x_i)^T G^{-1} \overline{k}(x_j),
\end{array}
\end{equation*}
where the elements of the usual Gram Matrix, $K$ is represented by $[K]_{ij} = k(x_i,x_j)$. Let us represent $\overline{k}(x_i)^T G^{-1} \overline{k}(x_j)$ in matrix form.  Notate\\
\begin{equation*}
    L = \begin{bmatrix}
    k(x_1,0) & k(x_1,1)\\
    \vdots & \vdots\\
    k(x_n, 0) & k(x_n, 1) 
\end{bmatrix}= \begin{bmatrix}
        \overline{k}^T(x_1)\\
        \vdots\\
        \overline{k}^T(x_n)
    \end{bmatrix},
\end{equation*}
then
\begin{equation*} \begin{array}{ll}
    \begin{bmatrix}
        \overline{k}^T(x_1)\\
        \vdots\\
        \overline{k}^T(x_n)
    \end{bmatrix} G^{-1} 
    \begin{bmatrix}
        \overline{k}(x_1) & \cdots & \overline{k} (x_n)\end{bmatrix}
        & = \begin{bmatrix}
            \overline{\alpha}^T(x_1) \\
            \vdots\\
            \overline{\alpha}^T (x_n)
        \end{bmatrix}    \begin{bmatrix}
        \overline{k}(x_1) & \cdots & \overline{k} (x_n)\end{bmatrix}\\
        & = \begin{bmatrix}
        \overline{\alpha}^T(x_1)\overline{k}(x_1) & \cdots &   \overline{\alpha}^T(x_1)\overline{k}(x_n)  \\
        \vdots & \ddots & \vdots \\
        \overline{\alpha}^T(x_n)\overline{k}(x_1) & \cdots &   \overline{\alpha}^T(x_n)\overline{k}(x_n)  
        \end{bmatrix}
        \end{array}.
\end{equation*}
Hence, 
\begin{equation*}
    K_0 = K - LG^{-1}L^T
\end{equation*}
 In the case of an explicit kernel, we can write $k(x,y) = \gamma(x)^T\gamma(y)$, where $\gamma: \mathcal{X} \rightarrow \mathbb{R}^p$ is a feature map. We compute:
\begin{equation*}
    \begin{array}{ll}
        k_0(x,y)
       & = \gamma(x)^T\gamma(y) - \begin{bmatrix}
            \gamma(x)^T \gamma(0) , \gamma(x)^T \gamma(0)
        \end{bmatrix} G^{-1} \begin{bmatrix}
            \gamma(y)^T \gamma(0) , \gamma(y)^T \gamma(1)
        \end{bmatrix}^T  \\
       & = \gamma(x)^T\gamma(y) - \begin{bmatrix}
            \gamma(x)^T \gamma(0) , \gamma(x)^T \gamma(1)
        \end{bmatrix} G^{-1} \begin{bmatrix}
            \gamma(0)^T \gamma(y) , \gamma(1)^T \gamma(y)
        \end{bmatrix}^T \\
     & = \gamma(x)^T\gamma(y) 
      - \gamma(x)^T\begin{bmatrix}
             \gamma(0) , \gamma(1)
        \end{bmatrix} G^{-1} \begin{bmatrix}
            \gamma(0), \gamma(1)
        \end{bmatrix}^T \gamma(y) \\
    & = \gamma(x)^T \left[ I  
      - \begin{bmatrix}
             \gamma(0) , \gamma(1)
        \end{bmatrix} G^{-1} \begin{bmatrix}
            \gamma(0), \gamma(1)
        \end{bmatrix}^T\right] \gamma(y). \\
    \end{array}
\end{equation*}
Let us define $\Gamma = \begin{bmatrix}
    \gamma(0)& \gamma(1)
\end{bmatrix} \in \mathbb{R}^{p\times2}$. where
\begin{equation*}
\begin{array}{ll}
G & = \Gamma^T \Gamma \\
& = \begin{bmatrix}
    \gamma(0) \\ \gamma(1)
\end{bmatrix}
\begin{bmatrix}
    \gamma(0) & \gamma(1)
\end{bmatrix} \\
&=\begin{bmatrix}
    \gamma(0)^T \gamma(0) & \gamma(0)^T \gamma(1)\\
    \gamma(1)^T\gamma(0) & \gamma(1)^T \gamma(1)
\end{bmatrix} \\
&= \begin{bmatrix}
    k(0,0) & k(0,1)\\
    k(1,0) & k(1,1)
\end{bmatrix}.
\end{array}
\end{equation*}
Then we can write
\begin{equation*}
    \begin{array}{ll}
     k_0(x,y)
    & = \gamma(x)^T \left[ I  
      - \begin{bmatrix}
             \gamma(0) , \gamma(1)
        \end{bmatrix} G^{-1} \begin{bmatrix}
            \gamma(0), \gamma(1)
        \end{bmatrix}^T\right] \gamma(y) \\
        & = \gamma(x)^T[I - \Gamma G^{-1} \Gamma^T ] \gamma(y).
    \end{array}
\end{equation*}
Let $P = \Gamma G^{-1} \Gamma^T$. Then $P$ satisfies the properties of a projection matrix onto the column space of $\Gamma$. This follows from the standard projection formula $P = \Gamma(\Gamma^T\Gamma)^{-1}\Gamma^T$ when projecting on the column space of a matrix $\Gamma$. $P$ is symmetric and idempotent:
\begin{equation*}
   \begin{array}{ll}
    P & =\Gamma G^{-1}\Gamma^T = (\Gamma G^{-1}\Gamma^T)^T \\
    & = P^T   \\
    P^2  &=(\Gamma G^{-1}\Gamma^T )^2 \\
    &=\Gamma G^{-1}\Gamma^T \Gamma G^{-1}\Gamma^T \\
    & = \Gamma G^{-1} G G^{-1} \Gamma^T\\
    & = \Gamma  G^{-1} \Gamma^T\\
    & = P.
   \end{array} 
\end{equation*}
Hence $P$ is an orthogonal projection matrix.  Thus, the matrix
\begin{equation*}
    A = I-P = I - \Gamma G^{-1} \Gamma^T,
\end{equation*}
represents the projection onto the orthogonal complement of the space spanned by $\gamma(0)$ and $\gamma(1)$. 
Since $P$ projects onto a 2-dimensional subspace, rank$(P) = 2$. Then $A = I-P$ is the complement of $P$, hence by the Rank-Nullity Theorem
\begin{equation*}
  \mathrm{rank}(A) + \mathrm{nullity}(A) = p,
\end{equation*}
and since $\mathrm{nullity}(A)=\mathrm{rank}(P)=2$, we get $\mathrm{rank}(A)=p-2$.
Now we can represent the modified kernel as:
\begin{equation*}
    k_0(x,y) = \gamma(x)^T A\gamma(y).
\end{equation*}
By spectral decomposition, $A = U\Lambda U^T$ where $U^TU = I$, $U = (u_1, \dots, u_{p-2})$, the matrix of eigenvectors of $A$ and $\Lambda$ the matrix of positive $p-2$ eigenvalues of $A$. Then,
\begin{equation*}
    \gamma_0(x) = \Lambda^{1/2} U^T \gamma(x),
\end{equation*}
is the feature vector of the explicit kernel $k_0$ with dimension $p-2$.

\end{document}